\documentclass[utf8]{FrontiersinVancouver} 

\usepackage{url,hyperref,lineno,microtype,subcaption}
\usepackage[onehalfspacing]{setspace}

\usepackage{subfig}
\usepackage{booktabs}
\usepackage{amsmath}
\usepackage{amsfonts}
\usepackage{amssymb}
\usepackage{graphicx}
\usepackage{subcaption}
\usepackage{algorithm}
\usepackage {algpseudocode}
\usepackage{algorithmicx}
\usepackage{algcompatible}
\usepackage{colortbl}
\usepackage{soul}
\usepackage{placeins} 

\nolinenumbers

\def\keyFont{\fontsize{8}{11}\helveticabold }
\def\firstAuthorLast{de Tinguy {et~al.}} 
\def\Authors{Daria de Tinguy\,$^{1,*}$, Tim Verbelen\,$^{2}$ and Bart Dhoedt\,$^{1}$}
\def\Address{$^{1}$Ghent University, Ghent, Belgium  \\
$^{2}$Verses, Los Angeles, California, USA }
\def\corrAuthor{Daria de Tinguy}

\def\corrEmail{daria.detinguy at ugent.be}

\begin{document}
\onecolumn
\firstpage{1}

\title[Learning Dynamic Cognitive Map with Autonomous Navigation]{Learning Dynamic Cognitive Map with Autonomous Navigation} 

\author[\firstAuthorLast ]{\Authors} 
\address{} 
\correspondance{} 

\extraAuth{}

\maketitle

\begin{abstract}

Inspired by animal navigation strategies, we introduce a novel computational model to navigate and map a space rooted in biologically inspired principles. Animals exhibit extraordinary navigation prowess, harnessing memory, imagination, and strategic decision-making to traverse complex and aliased environments adeptly. Our model aims to replicate these capabilities by incorporating a dynamically expanding cognitive map over predicted poses within an Active Inference framework, enhancing our agent's generative model plasticity to novelty and environmental changes. Through structure learning and active inference navigation, our model demonstrates efficient exploration and exploitation, dynamically expanding its model capacity in response to anticipated novel un-visited locations and updating the map given new evidence contradicting previous beliefs. Comparative analyses in mini-grid environments with the Clone-Structured Cognitive Graph model (CSCG), which shares similar objectives, highlight our model's ability to rapidly learn environmental structures within a single episode, with minimal navigation overlap. Our model achieves this without prior knowledge of observation and world dimensions, underscoring its robustness and efficacy in navigating intricate environments.

\tiny
 \keyFont{ \section{Keywords:} autonomous navigation; active inference; cognitive map; structure learning; dynamic mapping; knowledge learning} 
\end{abstract}

\section{Introduction}

Humans effortlessly discern their position in space, plan their next move, and rapidly grasp the layout of their surroundings~\cite{few_one_shot_learning, mice_in_labyrith} when faced with ambiguous sensory input~\cite{Human_rodent_spatial_rep}. Replicating these abilities in autonomous artificial agents is a significant challenge, requiring robust sensory systems, efficient memory management, and sophisticated decision-making algorithms.  Unlike humans, artificial agents lack inherent cognitive abilities and adaptive learning mechanisms, particularly when confronted with aliased observations, where sensory inputs are ambiguous or misleading~\cite{SLAM_alias}. 

To replicate human navigational abilities, an agent must capture the dynamic spatial layout of the environment, localise itself and predict the consequences of its actions. 
Most attempts to achieve this combine those fundamental elements in SLAM algorithms (Simultaneous Localisation and Mapping), often based on Euclidian maps~\cite{orb_slam3,survey_slam}. However, these methods require substantial memory as the world expands. Other strategies involve deep learning models, which depend on large datasets and struggle to adapt to unexpected events not encountered during training~\cite{exp_learning_survey}. A more efficient alternative lies in cognitive graphs or maps and learning a mental representation of the world from partial observations~\cite{World_Model, cscg_structuring_knowledge}, creating a symbolic structure of the environment~\cite{humans-cognitive-map,world_model_and_inference}. Cognitive graphs, by definition, represent a "mental understanding of an environment" derived from contextual cues like spatial relationships~\cite{dic_cognitive_map}. Alongside this structure, the ability to imagine the outcomes of actions enables more reliable navigation decisions based on preferences~\cite{nav_aif, human_exploration}.

Our approach integrates those biological capabilities into a unified model. Using visual observations and proprioception~\cite{grid_cell_nav}, we construct a cognitive map through a generative model, enabling navigation with an Active Inference (AIF) framework. This model links states by incorporating observations and positions through transitions, as illustrated in Figure~\ref{img:from_env_to_topo}b), showing the processed observation of the agent and c) presenting the resulting cognitive graph.
\begin{figure}[!hbt]
\setcounter{figure}{0}
    \centering
    \includegraphics[width=12cm]{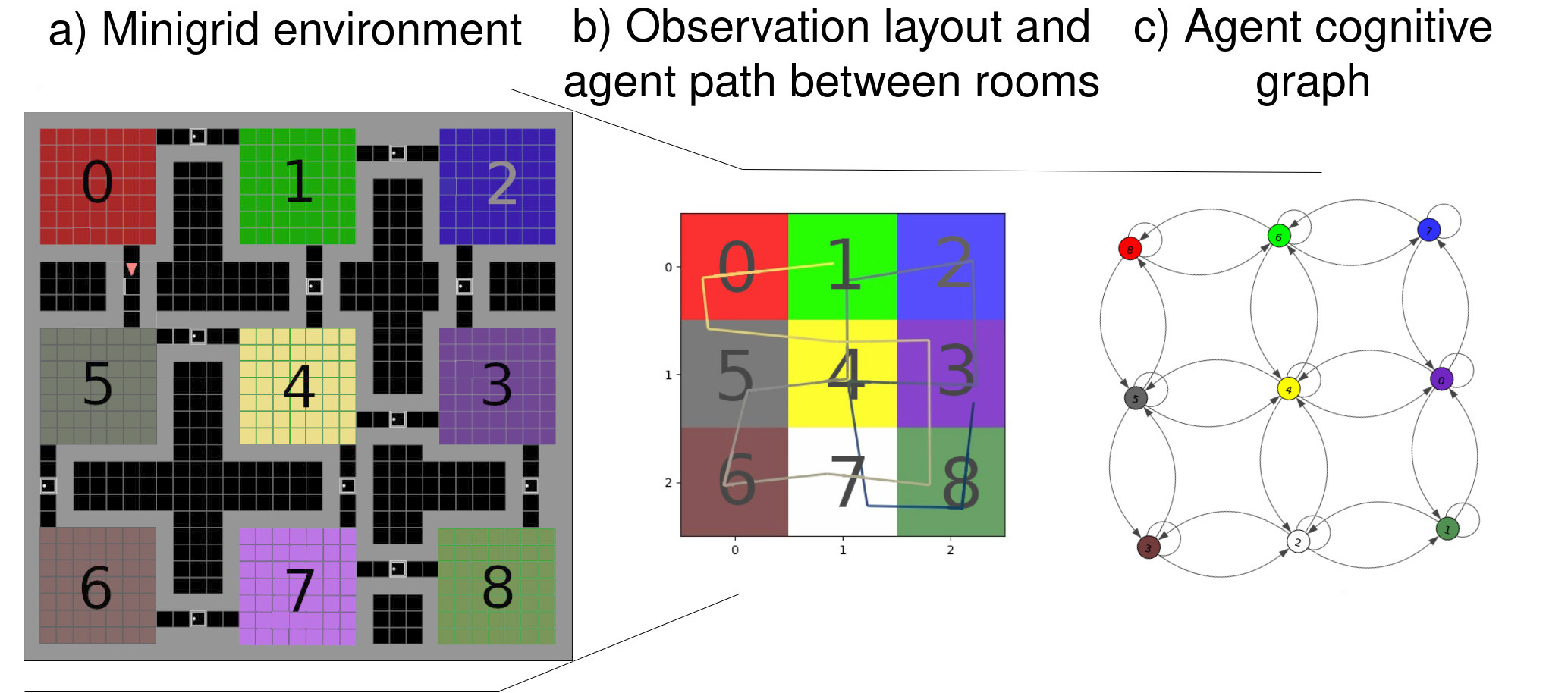}
    \caption{ a) From a full 3 by 3 rooms mini-grid environment~\cite{gym_minigrid, ours_2024} to b) rooms observation layout as perceived by the agent and the path it has taken between rooms - composed of a line from black to white-, c) shows the agent final internal topological graph (cognitive graph) linking all the locations between them.}
    \label{img:from_env_to_topo}
    \vspace{-0mm}
\end{figure}
Using Bayesian inference, the model predicts future states and positions, growing its cognitive map by forming prior beliefs about un-visited locations. As new observations are made, the agent updates its internal model, dynamically refining its representation of the environment~\cite{world_model_and_inference}. This continual adjustment allows the agent to effectively navigate complex environments by anticipating and learning from uncharted areas~\cite{build_cogn_map}. Our internal positioning system draws inspiration from the neural positioning system found in rodents and primates, aiding in self-localisation and providing an intrinsic metric for measuring distance and relative direction between locations~\cite{grid_cell_nav, Human_rodent_spatial_rep,grid_place_cells_clustered_envs}.

To achieve goal-directed navigation and exploration, we employ AIF  to model the agent's intrinsic behaviour in a biologically plausible way. Unlike methods relying on pre-training for specific environments, our approach introduces a navigation and dynamic cognitive map growing based on the Free Energy (FE) principle. This map is inspired by mechanisms observed in animals, such as border cells for obstacle detection~\cite{border_cells} and the visual cortex for visual perception. It continuously expands by predicting new observations and adapting dynamically to changing environmental structures.

This work aims to develop an autonomous agent that can determine where it is, decide where to navigate, and learn the structure of complex, unknown environments without prior training, mimicking the adaptability and spatial awareness observed in biological organisms. Traditional exploration approaches and deep learning models struggle in dynamic settings, requiring extensive memory and pre-collected datasets to predict future settings, or they face difficulties in adapting to untrained situations. The challenge is to design a model that allows agents to autonomously build, update, and expand an internal map based on current sensory data and past beliefs, efficiently managing ambiguous observations (such as aliased states) and responding flexibly to unexpected environmental changes.

Our contribution to this problem encompasses several key aspects:
\begin{itemize}
    \item Proposing a novel dynamic cognitive mapping approach that allows agents to predict and extend their internal map over imagined trajectories, enabling anticipatory navigation and rapid adaptation to new environments.  
    \item  Developing a navigation model that operates without pre-training or prior exposure, allowing the agent to successfully explore and make decisions in unfamiliar environments. 
    \item Proposing a flexible navigation behaviour fully explicit by relying upon the AIF framework.
    
    \item Outperforming in environmental learning and decision-making efficiency the Clone-Structured Cognitive Graph (CSCG) model~\cite{cscg_pres}, a prominent model for cognitive map representation~\cite{cscg_pres}.

    \item Showcasing robust adaptability, where the model responds seamlessly to dynamic environmental changes, replicating rat maze-like scenarios, thus emphasising its practical application in flexible, real-world navigation tasks. 

    \item Incorporating biologically inspired processes like border cells and visual cortex perception, our agent's navigation strategy is theoretically grounded and scalable to more realistic settings. 

\end{itemize}



\section{Related work}

As agents navigate their surroundings, they connect observations to construct an internal map or graph of the environment. This intuitive decision-making process is propelled by incentives such as food, safety, or exploration, rapidly guiding agents toward their objective \cite{mice_in_labyrith}.

\textbf{Motion planning}




Navigation tasks are often categorised into two primary scenarios based on the agent's familiarity with the environment and its objectives. In scenarios where the environment is partially known and the agent is already aware of its destination, the primary focus is achieving efficient retrieval and executing actions reliably. This entails leveraging existing knowledge about the environment's structure and landmarks to navigate swiftly and accurately towards the predetermined goal~\cite{exp_learning_survey}. Typically, this task centers around solving the motion planning problem: "How to move from point A to B?" Methods such as those in~\cite{log-MPPI, MPPI} propose enhanced versions of Model Predictive Path Integral (MPPI) navigation, which calculates the optimal sequence of control actions by simulating multiple future trajectories and selecting the one minimising a cost function. They allow real-time adaptation to obstacles in dynamic environments by continuously updating their navigation based on the most recent state and cost estimates of the control trajectory. However, these methods often rely on precise localisation within a local map, require substantial computational capacity, and may be vulnerable to sensor failures.

\textbf{Goal identification}

Our current model is not focusing specifically on solving motion planning, but on determining where the agent is and where it should go based on available information. Unlike traditional models, it does not require precise or absolute knowledge of its position; instead, it relies on its internally inferred belief about its location and considers obstacles relative to this position rather than to a global map. As a result, sensor failures are less critical, provided the agent can adapt to situations with either no sensory input or multiple inputs. In this work, the agent is expected to receive visual observations and detect obstacles through a system analogous to LiDAR. Failure of these sensors, however, would halt navigation.

When the agent must determine both its location and intended direction in addition to planning its movement, the task becomes significantly more complex, especially in unknown environments.
The agent must engage in map building and employ a form of reasoning to effectively operate and navigate through unfamiliar surroundings~\cite{neural_slam, rapid_task_solving}. This involves dynamically constructing a cognitive map of the environment, integrating sensory information, and adapting behaviour based on learned spatial relationships and environmental cues~\cite{learn_to_nav,ghost_node_work}. Works relying on Gaussian processes such as ~\cite{explo_gaussian_process} unify navigation, mapping, and exploration, reducing redundancy and improving exploration efficiency, however, it is weak to featureless environments, over-reliant on a correct position estimation and are weak to sudden displacements.
Our research addresses the challenges posed by determining where it should go and how it should do it by proposing an agent capable of seeking information gain and reaching desired observations in both familiar and unfamiliar environments. Our approach ensures robust performance even in scenarios involving potential disruptions, such as kidnapping, repeated observations or environmental modifications, without requiring any environment-specific pre-training. Through this framework, we seek to advance the capabilities of fully autonomous agents in navigating and solving tasks in diverse and dynamic environments.

\textbf{Spatial representation}

Spatial representation plays a significant role in robot navigation and boasts a rich history, offering diverse approaches with their own set of advantages and challenges~\cite{survey_slam}. While many SLAM systems rely on metric maps to navigate, which provide precise spatial information~\cite{SLAM_alias, orb_slam3}, there is a growing interest in topological mapping~\cite{ghost_node_explo, explo_gaussian_process} due to its biological plausibility and lower computational memory requirements. 
Cognitive maps are mental representations of spatial knowledge that explain how agents navigate and apprehend their environment~\cite{cscg_pres}. A cognitive map encompasses the layout of physical spaces~\cite{cscg_structuring_knowledge}, landmarks, distances between locations, and the relationships among different elements within the environment~\cite{humans-mapping,humans-cognitive-map}.
Models like the Clone-Structured Cognitive Graph (CSCG)~\cite{cscg_abstraction} and Transformer representations~\cite{cscg_transformers} create cognitive maps (usually represented as topological graphs) using partial observations, offering reusable, and flexible representations of the environment. However, these models often entail significant training time, typically involving fixed policies or random motions and require a statically defined cognitive map dimension. 

In biological systems, such as animals, the hippocampus plays a crucial role in managing episodic memory, spatial reasoning, and rapid learning \cite{hippo_pred_map} while structured knowledge about the environment is gradually acquired by the neocortex~\cite{Human_rodent_spatial_rep, cscg_space_latent_seq}. This enables remarkable adaptability and efficiency in navigation, with animals often requiring minimal instances to learn and navigate complex environments \cite{few_one_shot_learning}. They leverage cognitive mapping strategies to adapt to changes, swiftly grasp the layout of their surroundings, and efficiently return to previously visited places.

Drawing inspiration from these natural mechanisms, the compact cognitive map~\cite{brain_cognitive_map_SLAM} proposes an extendable internal map based on movement information (how far the agent is from any past location) and current visual recognition. 
Our system shares this goal of remembering significant observations, forming a spatial representation, and resolving ambiguity through contextual cues~\cite{humans-hierarchic-plan-clusters}. Furthermore, we extend its adaptability by proactively forecasting the extension of the internal cognitive map before the experience of new spatial information. This anticipatory capability empowers the system to predict potential action outcomes in unobserved areas, enhancing its ability to navigate in unknown territory. This proactive approach aligns more closely with the efficient decision-making processes observed in biological agents, allowing for rapid adaptation and effective navigation in diverse environments.

\textbf{Active Inference}
 
Animals navigate their environments adeptly by combining sensory inputs with proprioception~\cite{Human_rodent_spatial_rep}. This enables them to avoid being misled by repeated evidence, a phenomenon known as aliasing. Such navigation strategies can be elucidated through Active Inference~\cite{nav_aif}. Active Inference applied to navigation, entails the continual refinement of internal models based on sensory feedback, facilitating adaptive and effective decision-making within the environment. Serving as a normative framework, AIF elucidates cognitive processing and brain dynamics in biological organisms~\cite{life_friston,AIF_book}. It posits that both action and perception aim to minimise an agent's Free Energy, acting as an upper limit to surprise. Central to active inference are generative models, encapsulating causal relationships among observable outcomes, agent actions, and hidden environmental states. These environmental states remain 'hidden' as they are shielded from the agent's internal states by a Markov blanket~\cite{World_Model}. Leveraging partial observations, the agent constructs its own beliefs regarding hidden states, enabling action selection and subsequent observation to refine its beliefs relying on the Partially Observable Markov Decision Model (POMDP)~\cite{AIF_learning}.

At the heart of decision-making and adaptive behaviour lies a delicate balance between exploitation and exploration~\cite{curiosity_exploitative}. Exploitation involves selecting the most valuable option based on existing beliefs about the world, while exploration entails choosing options to learn and understand the environment~\cite{world_model_and_inference}. Recent behavioural studies indicate that humans engage in a combination of both random and goal-directed exploration strategies~\cite{human_exploration}. In our model, policies are chosen stochastically using the concept of FE. Regardless of the strategy employed, the agent attempts to minimise surprise by formulating policies that increase the likelihood of encountering preferred states, whether it is a specific observation (e.g. food) or the curiosity to comprehend the environment's structure~\cite{curiosity_exploitative}. This approach facilitates active learning by swiftly diminishing uncertainty regarding model parameters and enhancing knowledge acquisition about unknown contingencies. It also regulates model parameter updates in response to new evidence in a changing environment.
Typically this decision-making relies on priors over the current world structure and past observed outcomes to predict the next policies, usually leaning on static structures. Past research such as Bayesian model reduction techniques~\cite{bayesian_model_reduc}, or selection mechanisms~\cite{surpervised_struct_learning}, aim to expand models to accommodate emerging patterns of observation, relying on past and current observations. We propose to go one step further by extending our cognitive map over predicted outcomes, allowing the agent to reason over future un-visited states. 

All those mechanisms lead an agent to efficiently explore or forage autonomously in any structured environment, even if it changes. It is learning as it goes in a few-shot or one-shot learning as would mice in a labyrinth~\cite{mice_in_labyrith} relying on predictions and observations.

\section{Method}

\begin{figure}[!htb]
    \centering
    \includegraphics[width=3.5cm]{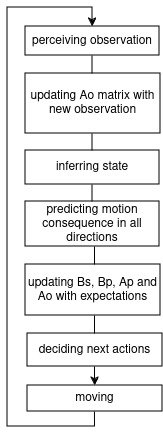}
    \caption{This block diagram outlines the agent's decision-making and mapping process. The agent begins by perceiving an observation, which updates the $A_o$ matrix—representing the observation model—by adjusting its dimensions to incorporate new sensory data. The agent then infers its current state based on this updated model. Next, it predicts the outcomes of possible motions in all directions, enabling it to update the model in relevant directions. The transition models, $B_s$, $B_p$, $A_p$ and $A_o$, are updated with anticipated new transitions (to new state or same state if obstacle), where $B_p$ represents probabilistic transitions between positions due to motion, $B_s$ captures state transition based on imagined spatial structure, and $A_p$ encoding the probability of a position given a state. Using this information, the agent decides on its next action and executes it by moving, repeating the process. }
    \label{img:step_diagram}
\end{figure}

The method section describes our approach to dynamic cognitive mapping and navigation in unfamiliar environments. Figure~\ref{img:step_diagram} provides a step-by-step overview of our model's process, visualised as a flowchart. This includes observing the environment, updating internal matrices with new information, inferring the current state, predicting possible motions, and selecting the next action based on updated beliefs.

To demonstrate the agent's navigation abilities, we tested it in a series of mini-grid environments, as illustrated in figure~\ref{img:from_env_to_topo}, where each map comprises rooms with a consistent floor colour connected by randomly positioned corridors separated by closed doors, all the map layouts used in this work are presented in Appendix~\ref{app:envs} and will be further detailed in section~\ref{results}. 

Our agent starts exploring without prior knowledge of the environment's dimensions or potential observations. It starts by inferring the initial state and pose from the initial observation and expands its model based on anticipating the outcomes of moving in the four cardinal directions. It accounts for potential unexplored adjacent areas before formulating decision-making policies, based on the model's existing beliefs. In Figure~\ref{img:from_env_to_topo}a), the agent's traversal of an environment is depicted, with each room encoded as a state by the model and the visual observation being the colour of the room's floor. Motion between rooms through a door suggests a transition between states, while walls are recognised as obstacles through RGB observations. The presented inference mechanism is expected to operate at the highest level of abstraction within a hierarchical framework such as~\cite{ours_2024}, where lower layers handle motion, the observation process and detect the presence of obstacles, akin to the output expected from, respectively, our motor cortex, visual cortex and border cells mechanism~\cite{border_cells}.  

Figure~\ref{img:from_env_to_topo}b) illustrates our agent navigation through processed observations, in the case of our mini-grid environments, the lower levels of the hierarchical model summarise the room information to a single colour (i.e., the floor colour) and obstacle detection (i.e., walls or doors) fed to the highest level of abstraction (i.e., presented model). The presented model generates an internal topological map given motions between rooms such as the final map shown in Figure~\ref{img:from_env_to_topo}c). Each state contains information about the room‘s floor colour and its inferred position. 


\subsection{State Inference}

\begin{figure}[!htb]
    \centering
    \includegraphics[width=6cm]{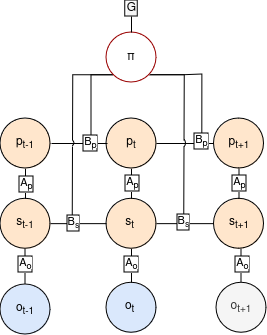}
    \caption{Factor graph of the POMDP in our generative model, showing transitions from the past to the present (up to time-step $t$) and extending into the future (time-step $t+1$). Past observations are marked in blue, indicating they are known. In the future steps, actions follow a policy $\pi$ influencing the new states and position in orange and new predictions in grey. The position at time $t$, $p_t$, is determined by the policy and the prior position $p_{t-1}$, while the current state $s_t$ is inferred from the observation $o_t$, the position $p_t$, and the previous state $s_{t-1}$.  Transitions between states are ruled by the $B$ matrices, which define how prior conditions contribute to the current one considering taken actions. $A$ matrices represent conditional probabilities of the quantities they connect.}
    \label{img:pomdp}
\end{figure}

Before each step, the agent engages in state inference by integrating the latest observation and internal positioning, following the POMDP graph depicted in Figure~\ref{img:pomdp}, where the current state $s_t$ and position $p_t$ are inferred based on the previous state $s_{t-1}$, position $p_{t-1}$ and action $a_{t-1}$ leading to the current observation $o_t$. The generative model capturing this process is described by Equation~\eqref{eq1}, where the joint probability distribution over time sequences of states, observations, and actions is formulated. Tildes~(~$\tilde{}$~) denote sequences over time. 

\begin{equation} 
P(\tilde{o}, \tilde{s}, \tilde{p} ,\tilde{a}) = P(o_0| s_{0})P(s_0)P(p_0) \prod_{t=1}^\tau 
P(o_t| s_{t})P(s_t, p_t|s_{t-1},p_{t-1},a_{t-1})  
\label{eq1}
\end{equation} 
The inference process operates within an AIF framework, where sensory inputs and prior beliefs are integrated to infer the agent's current state.

Due to the posterior distribution over a state becoming intractable in large state spaces, we use variational inference instead. This approach introduces an approximate posterior denoted as $Q(\tilde{s},\tilde{p}| \tilde{o}, \tilde{a})$ and presented in Equation~\eqref{eq2}~\cite{aif_step_by_step} .

\begin{equation} 
Q(\tilde{s},\tilde{p}| \tilde{o}, \tilde{a}) = Q(s_0, p_0| o_0) \prod_{t=1}^\tau
Q(s_t, p_t| s_{t-1},p_{t-1}, a_{t-1}, o_t )  
\label{eq2}
\end{equation}

The inference scheme defined in Equation~\eqref{eq2}, heavily relies on priors and observations to localise the agent within its environment.  

Therefore, we associate observations $o_t$, stemming from visual information, with our inferred position $p_t$, generated by the agent's proprioception. This improves state inference in an aliased environment, where the agent is simultaneously inferring its state and building its cognitive map. In the absence of prior information, the internal positioning $p$ is initialised at the start of exploration as an origin (i.e., a tuple (0,0)). It is then updated when the agent transitions between rooms (e.g., by passing through a door). 

When the agent is certain about its current state, it updates past beliefs about observations and transitions in response to environmental changes. This ensures rapid re-alignment between the agent's model and the evolving environment, within one or a few iterations, depending on the agent's confidence in its outdated beliefs. The specific learning mechanism of our model will be explained in section~\ref{sec:cogn_map_ext}.

In scenarios where the agent is unexpectedly relocated to an unfamiliar position, as tested in some of our experiments, its certainty about its location decreases, since its current state and previous actions no longer match its present observations. This drop in confidence occurs as the probability of being in a particular state—considering both position and observation—falls below a predefined certainty threshold (see Appendix~\ref{app:models_params} for exact values). When this threshold is crossed, the agent suspends reliance on its position information to infer state, focusing instead exclusively on observations to regain sufficient confidence over its state. The position and model updates are paused during this re-estimation of its location until the agent achieves adequate confidence in its state. This approach, guided by Free Energy minimisation, prioritises gathering information about its whereabouts, fostering informed decision-making in uncertain conditions and encouraging a resilient comprehension of its new state and position.

\subsection{Free Energy}
Typically, agents are assumed to minimise their variational free energy denoted $F$, which can serve as a metric to quantify the discrepancy between the joint distribution $P$ and the approximate posterior $Q$ as presented in Equation~\eqref{eq:F}~\cite{AIF_book}. This equation shows how adequate is the model to explain the past and current observations.

\begin{equation}
\begin{aligned}
F & = \mathbb{E}_{Q(\tilde{s},\tilde{p}|\tilde{a},\tilde{o})}[\log[Q(\tilde{s},\tilde{p}|\tilde{a},\tilde{o})] - \log[P( \tilde{s},\tilde{p},\tilde{a},\tilde{o})]] \\
 & = \underbrace{D_{KL}[Q(\tilde{s},\tilde{p}|\tilde{a},\tilde{o}))||P(\tilde{s},\tilde{p}|\tilde{a},\tilde{o})]}_\text{posterior approximation} -\underbrace{\log[ P(\tilde{o})]}_\text{log evidence}\\
 & = \underbrace{D_{KL}[Q(\tilde{s},\tilde{p}|\tilde{a},\tilde{o}))||P(\tilde{s},\tilde{p}|\tilde{a})]}_\text{complexity} - \underbrace{\mathbb{E}_{Q(\tilde{s},\tilde{p}|\tilde{a},\tilde{o})}[\log[P(\tilde{o}| \tilde{s})]}_\text{accuracy}
\end{aligned}
\label{eq:F}
\end{equation}

Active inference agents aim to minimise their free energy by engaging in three main processes: learning, perceiving, and planning. Learning involves optimising the model parameters, perceiving entails estimating the most likely state, and planning involves selecting the policy or action sequence that leads to the lowest expected free energy. Essentially, this means that the process implicates forming beliefs about hidden states offering a precise and concise explanation of observed outcomes while minimising complexity~\cite{life_friston}.  

While planning, we minimise the Expected Free Energy (EFE) instead, denoted $G$, indicating the agent's anticipated free energy after implementing a policy $\pi$. Unlike the variational free energy $F$, which focuses on current and past observations, the expected free energy incorporates future expected observations generated by the selected policy. 

To calculate this expected free energy $G(\pi)$ over each step $\tau$ of a policy we sum the expected free energy of each time-step. 

\begin{equation}
    G(\pi) = \sum_\tau G(\pi,\tau)
\end{equation}
\begin{equation}
\begin{aligned}
    G(\pi,\tau) =& \underbrace{\mathbb{E}_{Q(o_\tau,s_\tau, p_\tau|\pi)}[\log(Q(s_\tau,p_\tau|\pi) - \log(Q(s_\tau, p_\tau| o_\tau, \pi))]}_\text{information gain term} \\ 
    &- \underbrace{\mathbb{E}_{Q(o_\tau,s_\tau,p_\tau|\pi))}[\log(P(o_{\tau}))]}_\text{utility term}
\end{aligned}
\label{eq:planning_as_inf}
\end{equation}

The expected information gain quantifies the anticipated shift in the agent's belief over the state from the prior $Q(s_\tau|\pi)$ to the posterior $Q(s_\tau| o_\tau,\pi)$ when pursuing a particular policy. On the other hand, the utility term assesses the expected log probability of observing the preferred outcome under the chosen policy. This value intuitively measures the likelihood that the policy will guide the agent toward its prior preferences. Prior preferences are embedded within the agent's model as an objective or target state the agent should work toward. Usually given as a preferred state or observation. Free Energy indirectly encourages outcomes that align with its preferences or target states. This approach makes the utility term less about "reward" in the traditional sense of Reinforcement Learning and more about achieving coherence with the agent’s built-in preferences, balancing this with exploration (information gain).
The agent therefore chooses the optimal policy to follow among possible policies by applying: 
\begin{equation}
    P(\pi) = \sigma (-\gamma G(\pi))
\end{equation}
Where $\sigma$, the softmax function is tempered with a temperature parameter $\gamma$ converting the expected free energy of policies into a categorical distribution over policies. Playing with the temperature parameter alters the stochasticity of the navigation, the agent being more or less likely to choose the optimal policy rather than any other one. Details about the definition of policies can be found in Appendix~\ref{app:policies}.

\subsection{Active Inference and Cognitive Map Navigation}

When navigating within an environment, the agent uses AIF to continuously refine its knowledge of the world by updating its model parameters, more specifically, transition probabilities (how likely it is to move from one state to another) and likelihoods (how likely it is to observe certain features given a state). These quantities are updated, considering the actions taken by the agent and the observations gathered resulting from them. These updates help the agent to reduce the gap between its predicted and actual observations, effectively fine-tuning its internal model to match the real environment better. This approach allows the agent to anticipate future states better and select actions that will minimise future EFE, thus optimising its navigation strategy. Typically, this process assumes that the agent has some prior knowledge about the environment, such as its dimensions or the types of observations it might encounter~\cite{weird_HAIF,nav_aif}. This prior knowledge allows the agent to form expectations over observations or states, even though it might not know which observations correspond to which locations. For instance, the agent might expect to encounter walls, doors, or specific floor colours, but it does not know where exactly these will be.  

Identifying a model architecture with the right level of complexity to capture accurately the environment is problematic. Therefore, we take the approach of letting the model expand dynamically and explore as needed. Model expansion, however, requires a trigger indicating the need for an increase in model complexity.  

Some have expanded their model upon receiving new patterns of observations~\cite{ours_2024, surpervised_struct_learning}. When considering the case of exploring room structured mazes it implies the creation of new states only when the agent physically transitions to a new location/room. 

In an environment where multiple actions could have potentially created new states, only generating states upon actual observation means the agent will forget that previously visited rooms could have led to other unexplored spaces. Consequently, the generative model cannot leverage this knowledge to predict accurately the presence or absence of adjacent rooms from past visited rooms since these potential states have not been included in the model. 

Therefore, when the agent evaluates policies using EFE, it fails to accurately predict action consequences if it has not directly experienced those transitions. This short-sighted exploration strategy neglects the potential connections between visited and un-visited locations, leading to sub-optimal exploration. 

To illustrate this, consider our agent expanding its internal map only after directly observing the new rooms. Initially, the agent observes its current room and knows that doors may lead to other rooms, but it has no prior knowledge of what lies behind these doors. This uncertainty results in a high predicted information gain in the EFE. When the agent finally crosses a door and encounters a new observation, it generates a new state corresponding to this observation. However, the doors it did not explore are forgotten, as only really observed outcomes are considered to update the model. Thus, the agent will not consider these un-visited areas as offering significantly greater potential for information gain in its future planning. This leads to a slower exploration process, where the agent takes longer to visit all rooms because it fails to anticipate the existence of un-visited distant locations and ignores how to reach them. The model overlooks exploration opportunities that could help it select the most effective exploration strategy overall.  

To address this, our model learns to expand its internal map based on all available opportunities and grows based on predicted observations or states rather than solely on actual observations. This way, the agent retains awareness of all potential actions, including those it has not yet executed, allowing it to systematically explore the environment by considering both visited and un-visited locations. This strategy enhances the connectivity between different areas of the environment (enabling the agent to predict that different doors are likely to lead to the same room) and enables the agent to explore more efficiently, reducing the time required to map its surroundings fully. 

\begin{figure}[!htb]
    \centering
    \includegraphics[width=1\linewidth]{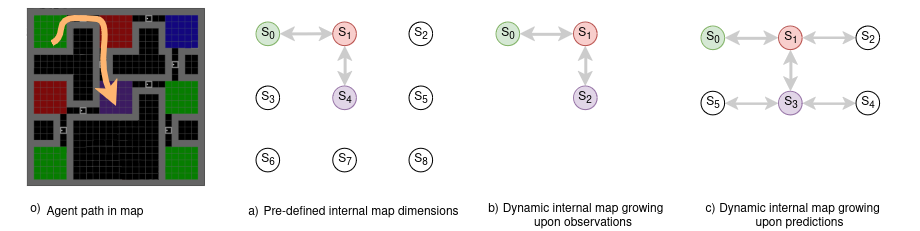}
    \caption{Example of an internal map layout based on initial dimensions, expansion strategy, and a given path in a maze -o) path in orange-. with observed states/rooms observations being the room floor. a) A static map expects 9 rooms/states but lacks connectivity and observation details. A dynamic map can expand indefinitely with b) new observations or c) predictions. a) Requires knowing the environment's size in advance, while b) does not foresee new rooms in un-visited past areas. c) Considers the possibility of every door leading to un-visited rooms.}
    \label{fig:int_map_nav}
\end{figure}

Figure~\ref{fig:int_map_nav} presents the three approaches: 
\begin{itemize}
    \item[] a) when we know the environment dimension (static cognitive map dimension),
    \item[] b) when the internal map grows upon receiving new observations, 
    \item[] c) when the internal map grows upon predicting new states yet to be explored.
\end{itemize}

the first plan o) presents an agent's motion in a room maze, the arrow in orange illustrates the path the agent follows from the green to the purple room. a) requires a prior over the environment, thus not being adapted to fully unknown mazes, b) is not considering the blue room in its model even though it knew it existed when in the red room ($S_1$), but since it chose to move toward the purple room, this knowledge is lost. c) Considers all the motions it could have taken to change rooms. Thus, it remembers that the blue room exists and that it has never been there, improving drastically exploration as it can plan to move there from anywhere in the maze.

\subsection{Cognitive Map Extension}
\label{sec:cogn_map_ext}
In this section, we present how the model learns the structure of the environment by expanding the cognitive map based on predictions, before obtaining actual observations, and updating beliefs upon given new evidence. 

Each time the agent moves, the model infers its current state and pose by integrating its motion and observations, subsequently updating the model parameters. In line with the Bayesian reduction model~\cite{bayesian_model_reduc}, if the observation likelihood $A_o$ fails to encompass the current observation, this indicates the novelty of the observation. As a result, the dimensionality of the observation likelihood is expanded to incorporate the new observation, and the model parameters $\theta$ are updated accordingly to reflect this extended observation likelihood.

The optimisation of beliefs regarding the generative model parameters $\theta$ involves minimising the Free Energy $F_\theta$, while accounting for prior beliefs and uncertainties related to both parameters and policies, as outlined in~\cite{AIF_book}: 
\begin{equation}
\begin{aligned}
    \theta =& (A_o,A_p,B_s)\\
    F_\theta =& \mathbb{E}_{Q(\pi,\theta)}[F(\pi,\theta)] + D_{KL}[Q(\theta)||P(\theta)] + D_{KL}[Q(\pi)||P(\pi)]
\end{aligned}
\label{eq:param_update}
\end{equation}

This optimisation in Equation~\ref{eq:param_update} balances the expected accuracy of the model's predictions and the necessity of maintaining coherence with prior beliefs. This ensures that the agent's learned representations and actions are both accurate and consistent with existing knowledge. The parameter set $\theta$ includes the following Markov matrices: the state transition $B_s=P(s_t|s_{t-1}, a_{t-1})$, the observation likelihood $A_o= P(o_t|s_t)$ and the position likelihood $A_p= P(p_t|s_t)$. 
The position transition matrix $B_p = P(p_t|p_{t-1}, a_{t-1})$ is not updated through Active Inference; instead, it is a deterministic metric in which the size grows as the agent explores, guided by the policy $\pi$. This indicates that the transition between two positions is determined by the given action and the previous position.

In contrast, Equation~\ref{eq:param_update} indicates that the state transition model $B_s$ is updated based on the transitions the agent experiences during exploration. Further details regarding the learning rates and matrix initialisation can be found in Appendix~\ref{app:models_params}.

Once the current model is fully updated, policy outcomes are predicted using Expected Free Energy (EFE) in the four cardinal directions from the agent's current state. Walls and doors are identified from RGB observations and incorporated to determine the feasible directions for room transitions, influencing the transition probabilities. Equation~\ref{eq:efe_wt_c} presents the EFE of a policy $\pi$ with collision observation $c$. The probability $P(c)$ is binary, meaning there either is or isn’t a collision, with probabilities of 1 or 0, respectively. The learning term of the equations shows how much we learn about the position likelihood considering an obstacle or not while the inference term evaluates the state $s_{t+1}$ and position $p_{t+1}$ given the collision observation $c_{t+1}$. It reflects how well the model infers the state and position based on that observation.

\begin{equation}
\begin{aligned}
G(\pi) &= \mathbb{E}_{Q_{\pi}} [\log Q(s_{t+1},p_{t+1}, A_p | \pi) - \log Q(s_{t+1},p_{t+1}, A_p | c_{t+1}, \pi) - \log P(c_{t+1})] \\
&= -\underbrace{\mathbb{E}_{Q_{\pi}} [\log Q(A_p |s_{t+1}, p_{t+1}, c_{t+1}, \pi) - \log Q(A_p |s_{t+1}, p_{t+1}, \pi)]}_\text{expected information gain (learning)}\\
&\quad -\underbrace{\mathbb{E}_{Q_{\pi}} [ \log Q(s_{t+1},p_{t+1} | c_{t+1}, \pi) - \log Q(s_{t+1},p_{t+1} | \pi) ]}_\text{expected information gain (inference)} \\
&\quad -\underbrace{\mathbb{E}_{Q_{\pi}} [ \log P(c_{t+1}) ]}_\text{expected risk of collision}
\end{aligned}
\label{eq:efe_wt_c}
\end{equation}

The transition model $B_p$ tracks possible transitions between positions. It determines the next position by incrementing the previous one based on the agent's motion. When calculating transition probabilities between positions, detected obstacles are taken into account. If an obstacle is detected in a given direction, the transition probability to that position is set to zero, and the predicted new position is disregarded. Conversely, if no obstacles are detected and the agent predicts a new position $p_{t+1}$ that does not match any previously known positions (i.e., the position has not been imagined before), the size of $B_p$ is expanded to include this newly discovered position. This approach enables the model to adapt and remember new positions as they are encountered. Considering any pose $p$, equation~\ref{eq:efe_over_param} represents the Expected Free Energy (EFE) of the prior over the position likelihood parameters, summed across all policies. The sigmoid function, $\sigma$, transforms the negative EFE into a probability value~\cite{surpervised_struct_learning}. This equation evaluates how effectively the current position likelihood $A_p$ explains the relationship between the state $s$, the pose $p$, and the collision observation $c$. 

\begin{equation}
\begin{aligned}
P(A_p) &= \sigma(-G) \\
G(A_p) &= \mathbb{E}_{Q_{A_p}} [\log P(p,s |A_p) - \log P(p, s |c, A_p) - \log P(c)] \\
&= -\underbrace{\mathbb{E}_{Q_{A_p}} [\log P(p, s|c, A_p) - \log P(s |p,A_p) - \log P(p|A_p)]}_\text{expected information gain} \\
&\quad  
- \underbrace{\mathbb{E}_{Q_{A_p}} [\log P(c)]}_\text{expected value}
\end{aligned}
\label{eq:efe_over_param}
\end{equation}

If the predicted next position $p_{t+1}$ does not correspond to any existing state in the model (considering the state probability conditioned on the position) the model recognises this position as corresponding to a new, unexplored state. Thus the model introduces a new state in its representation to account for this new place, expanding all probability matrices to accommodate the additional state dimension. This ensures that the state dimensions remain consistent across all matrices. The newly introduced state is assigned a high probability in the position likelihood matrix $A_p$, such that $P(s_{t+1}|p_{t+1})=1$, ensuring the model accurately reflects this new position in its state representation. 

The observation likelihood matrix $A_o$ assigns uniform probabilities across its distribution for un-visited states, reflecting the agent's limited knowledge about the unexplored states. Essentially, the agent cannot predict which observation to expect at a newly predicted location due to the lack of prior experience there.

The model first evaluates the likelihood of reaching the predicted position given a specific motion to calculate the transition probabilities between a current and predicted state based on position. If an obstacle is detected in the direction of this expected position, the transition probability is null, meaning the action results in the agent staying in the same state. Conversely, if no obstacle is detected, the transition likelihood between the current and predicted states is maximised. This state transition is determined by the position transition and the corresponding probability $P(s_t|p_t)$, ensuring that the position-based transition is correctly mapped to the state transition within the model. In both scenarios, the Dirichlet distribution of the transition matrix $B_s$ is updated similarly to how it is with experienced transitions, but with a lower learning rate. This lower learning rate accounts for the greater uncertainty associated with predicted transitions compared to those already experienced. Details about the learning rates can be found in Appendix~\ref{app:models_params}.

These unexplored states are highly attractive in the EFE framework because visiting them offers significant information gain.

\begin{figure}[!htb]
    \centering
    \includegraphics[width=6cm]{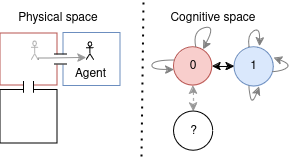}
    \caption{From a physical motion to the subsequent cognitive graph update. The first red room is initialised with current observation and predicted motions in the four directions. Going Down and Right holds new unknown states. By going right toward the blue room, the state is updated with a new blue observation and predicted motions in all directions, increasing the confidence in the red-blue room transition and defining it as bi-directional.}
    \label{img:ph_to_co}
    \vspace{5mm}
\end{figure}

The schematic depicted in Figure~\ref{img:ph_to_co} illustrates a straightforward process of updating the state from the physical world to the cognitive graph. Unknown states are weakly defined as existing but lacking any observations, resulting in weaker transition probabilities compared to discovered states through experimental transitions holding observations.

\section{Results}
\label{results}

Our experiments are designed to assess the efficacy of our model in constructing cognitive graphs by linking visited locations with anticipated ones, thereby enabling efficient exploration and goal achievement. 

We conducted comparative analyses with CSCG to assess several functionalities. These include learning spatial maps under aliased observations and decision-making strategies aimed at internal objectives, such as exploration or reaching specific observations. Those tests were done with and without prior environmental information (i.e. is the model familiar with the environment layout). Finally, we validated our agent's self-localisation capability following a kidnapping and re-localisation at a random place and how it re-plans after observing obstacles on its path.

The Clone-Structured Cognitive Graph (CSCG) is a graph-based cognitive architecture where nodes represent beliefs and edges are transitions between these states. Clones of these nodes allow the agent to maintain multiple hypotheses about its environment, enabling parallel exploration and evaluation of different action sequences and their outcomes. This structure supports dynamic and adaptive decision-making in aliased environments~\cite{cscg_pres}. More about the CSCG can be found in Appendix~\ref{app:cscg}.

The environments comprise interconnected rooms, ranging from configurations with 7 to 23 rooms. These arrangements vary from T-shaped layouts to 3 by 3, 4 by 4 room grids, donuts-shaped layouts and a reconstruction resembling a Tolman's maze, all with or without aliased room colours. Environments containing several rooms with the same floor colour are considered aliased and named such in our experiments. 

We consider both models (CSCG and ours) as part of a hierarchical framework~\cite{ours_2024}. Information is processed at the lower levels, generating for each room:
\begin{itemize}
    \item a single colour per room as observation
    \item spatial boundary information (i.e. expected risk of collision) indicating obstacles or possible ways out (doors)~\cite{border_cells}
\end{itemize}
The agents can move in the four cardinal directions or remain stationary in the current room. Moving towards a door leads to entering a new room while moving towards a wall does not alter the observation. Detailed observation layouts of the 8 environments used can be found in Appendix~\ref{app:envs}, with an example of the 3x3 rooms layout and observations illustrated in Figure \ref{img:from_env_to_topo}.

\subsection{Dynamic Extension}

By moving through space, gathering observations, and updating their internal beliefs, both our model and CSCG exhibit the ability to learn the underlying spatial map of the environment, akin to the navigational capabilities observed in animals~\cite{cscg_structuring_knowledge}. However, while a CSCG has a static cognitive map and learns through random exploration, our model starts with a smaller state dimension and makes informed decisions at each step based on its internal beliefs and preferences. This mechanism enhances the relevance of each movement and accelerates the learning process, mirroring the efficient navigation strategies seen in animals \cite{human_exploration}. 

\begin{table}[!htb]
\caption{The average number of steps required to fully learn the environments for each model.}
{\begin{tabular}{lllllll}
\begin{tabular}[c]{@{}l@{}}models\textbackslash\\ environments\end{tabular} & Oracle & ours & CSCG & \begin{tabular}[c]{@{}l@{}}CSCG\\ pose ob\end{tabular} & \begin{tabular}[c]{@{}l@{}}CSCG\\ random policy\end{tabular} & \begin{tabular}[c]{@{}l@{}}CSCG\\ pose ob\\ random policy\end{tabular} \\ \hline
3x3 & 11 & \textbf{14.4$\pm$2} & 231.9$\pm$59 & 221.3$\pm$35 & 214.6$\pm$44 & \multicolumn{1}{l}{205.7$\pm$22} \\ \hline
3x3\_alias & 11 & \textbf{18.1$\pm$2} & 376.3$\pm$35 & 201.5$\pm$20 & 338.8$\pm$30 & 210.4$\pm$22 \\ \hline
4x4 & 15 & \textbf{34.0$\pm$4} & 375.7$\pm$36 & 405.5$\pm$78 & 419.6$\pm$98 & 380.5$\pm$53 \\ \hline
4x4\_alias & 15 & \textbf{48.9$\pm$9} & 509.5$\pm$107 & 387.5$\pm$67 & 554.1$\pm$70 & 391.3$\pm$66 \\ \hline
T\_maze & 9 & \textbf{12.5$\pm$1} & 284.5$\pm$69 & 297.7$\pm$108 & 261.7$\pm$96 & 232.5$\pm$73 \\ \hline
T\_maze\_alias & 9.5 & \textbf{14.3$\pm$4} & 592.5$\pm$28 & 282.7$\pm$124 & 602.0$\pm$69 & 341.0$\pm$75 \\ \hline
donuts & 13 & \textbf{15.6$\pm$2} & 519.6$\pm$127 & 450.0$\pm$87 & 416.3$\pm$105 & 493.1$\pm$132
\end{tabular}}
\label{tab:explo_average}
\end{table}

\begin{figure}[!htb]
    \centering
    \includegraphics[width=5cm]{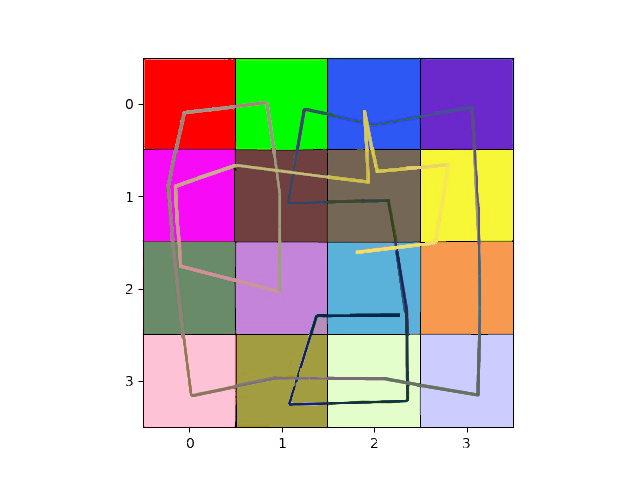}
    \caption{A path -from black to white- our agent takes to fully learn the environment, most rooms are accessed twice to learn the transitions between connected rooms fully.}
    \label{img:4x4_agent_path}
    \vspace{-0mm}
\end{figure}
\begin{figure}[!hb]
    \centering
    \includegraphics[width=10cm]{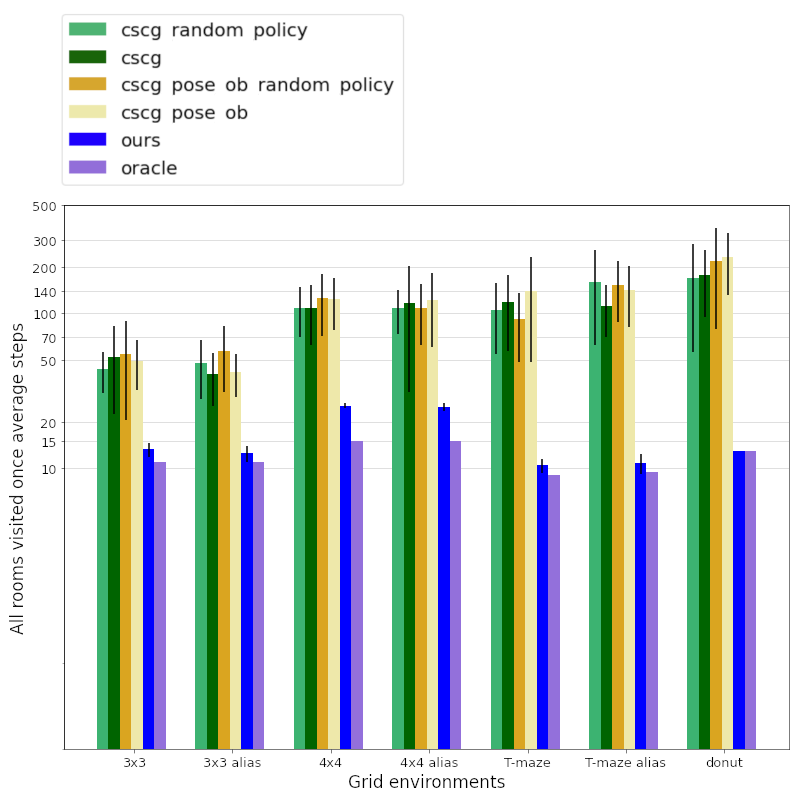}
\caption{The Figure displays the average number of steps required to discover all the rooms on a logarithmic scale, with the oracle serving as the benchmark for the minimum steps needed to visit all rooms once. Aliased rooms, marked by identical observations across different locations, present a challenge by potentially misleading the agent about its current position. Nonetheless, our agent has a meaningful pathway, discovering all rooms at least 30\% faster than CSCG.}
\label{img:average_steps_models_room_discovery_per_env}
\end{figure}
Both models were made to learn the connectivity between rooms. The agents spawned at random start locations and were tasked with seeking information gain from the environment -in other words, they were expected to learn the layout of the environment. Our model is led by information gain only as no preferred observations are given to the model, the utility term is silent during pure exploration. The results averaged over 20 tests per model and environment, are summarised in Table~\ref{tab:explo_average} and Figure~\ref{img:average_steps_models_room_discovery_per_env}. Our model demonstrates efficient exploration in all those environments. Exploration is considered complete when the internal belief regarding transitions between places aligns with the ground-truth transition matrix of the environment, with a minimum confidence of 60\% over all correct transitions. This level of confidence was deemed enough to guarantee a good understanding of the maze structure for all models. To ensure a fair comparison, we provided CSCG with either colour observations alone (referred to as 'CSCG' in our results) as in our proposed model, or colour-position pairs observations (termed 'CSCG pose ob' in our results). We considered providing ground-truth position and colour as an observation during exploration without prior as equivalent to our model's proprioceptive ability. Additionally, the CSCG can navigate randomly as in the original paper~\cite{cscg_pres} or by considering a localisation belief similar to our model, coupled with the CSCG Viterbi navigation module as was realised in \cite{Toon_integration}. Additional information about the CSCG architecture and training process can be found in appendix~\ref{app:cscg}. Two major differences highlight the capabilities of our model compared to the CSCG model: our model can expand its internal map dynamically, adapting to the environment without requiring a pre-defined dimension size, and it can reason over unexplored areas, whereas the CSCG model needs to encounter a new place to integrate that state into its internal map.

Table~\ref{tab:explo_average} shows the impact of this difference in design. We can compare the number of steps required by each agent to explore diverse environments, with the oracle, an A-star algorithm, defining the minimum number of steps needed to visit all rooms once. In the T-shaped maze, the oracle's number of steps is averaged considering their starting poses. Our model demonstrates significantly faster learning of all map structures than the CSCG algorithm. All CSCG models require a substantial number of steps to explore the environment, regardless of whether the observations are ambiguous or not. Neither random exploration nor Viterbi-based navigation significantly improves its performance. The decision-making process does not prioritise policies guiding the agent toward unexplored areas, limiting its overall effectiveness. Figure~\ref{img:4x4_agent_path} illustrates a typical path undertaken by our agent within a 4x4 observation environment. Despite a few instances of overlapping paths, our agent demonstrates a discerning approach, making informed decisions rather than resorting to random exploration. This strategic decision-making results in an average maximum number of steps of our model being about twenty times smaller than those of the CSCG. To delve deeper into the exploration process, we examine the first discovery of all rooms, as depicted in Figure~\ref{img:average_steps_models_room_discovery_per_env} on a logarithmic scale. Remarkably, our model's performance closely resembles the oracle's, suggesting a tendency to prioritise novelty exploration over confirming connections between rooms.
 Emphasising imagined beliefs leads to faster convergence of locations connectivity, however, it also carries the risk of forming false connections not substantiated by observation. To mitigate this risk, we opted to maintain state connections made out of imagined beliefs weaker than those formed through observation. 

\FloatBarrier

\subsection{Autonomous Localisation}

\setcounter{figure}{7}
\setcounter{subfigure}{0}
\begin{subfigure}
\setcounter{figure}{7}
\setcounter{subfigure}{0}
    \centering
    \begin{minipage}[b]{0.49\textwidth}
        \includegraphics[width=\linewidth]{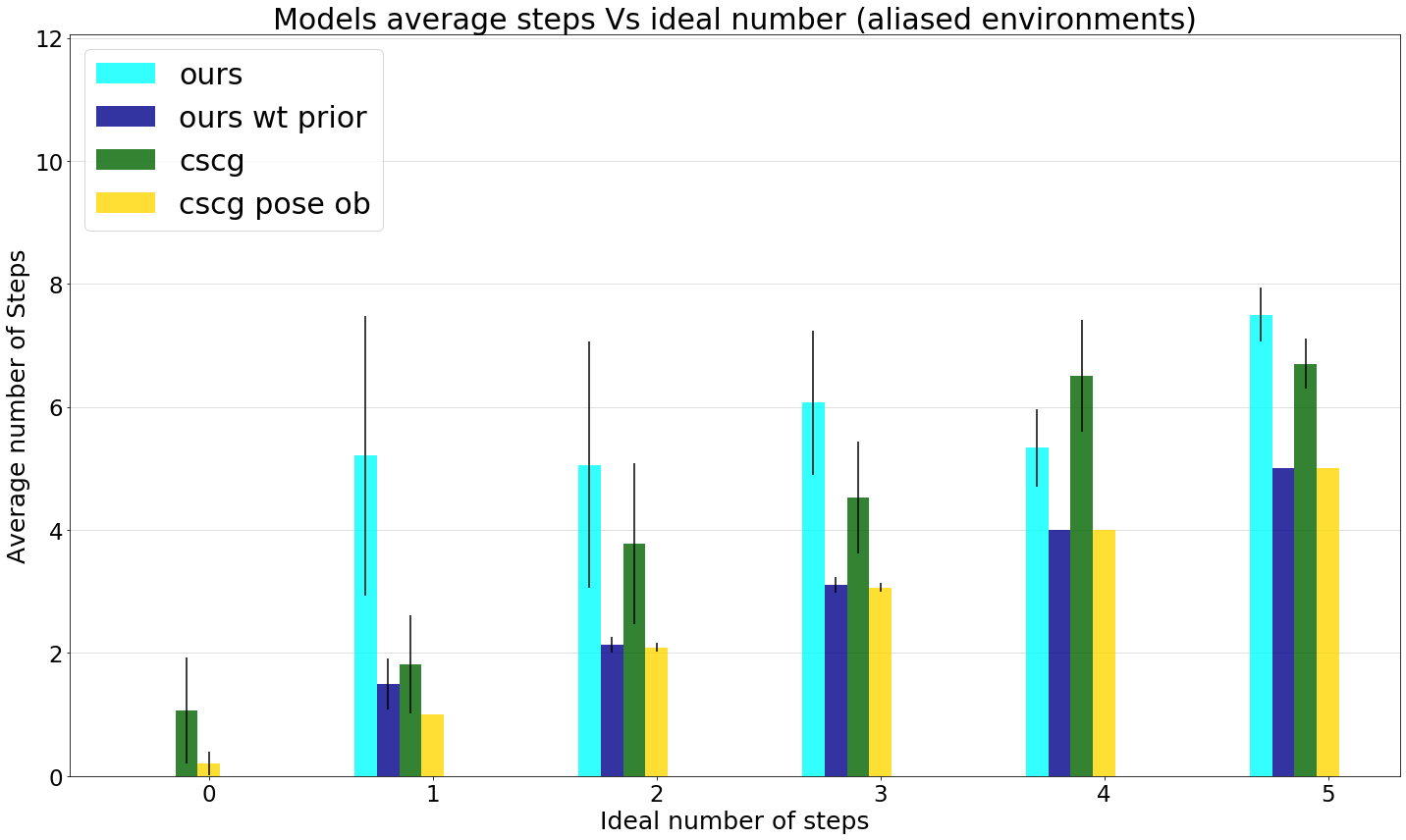}
        \caption{}
        \label{img:steps_goal_alias}
    \end{minipage}  
\setcounter{figure}{7}
\setcounter{subfigure}{1}
    \begin{minipage}[b]{0.49\textwidth}
        \includegraphics[width=\linewidth]{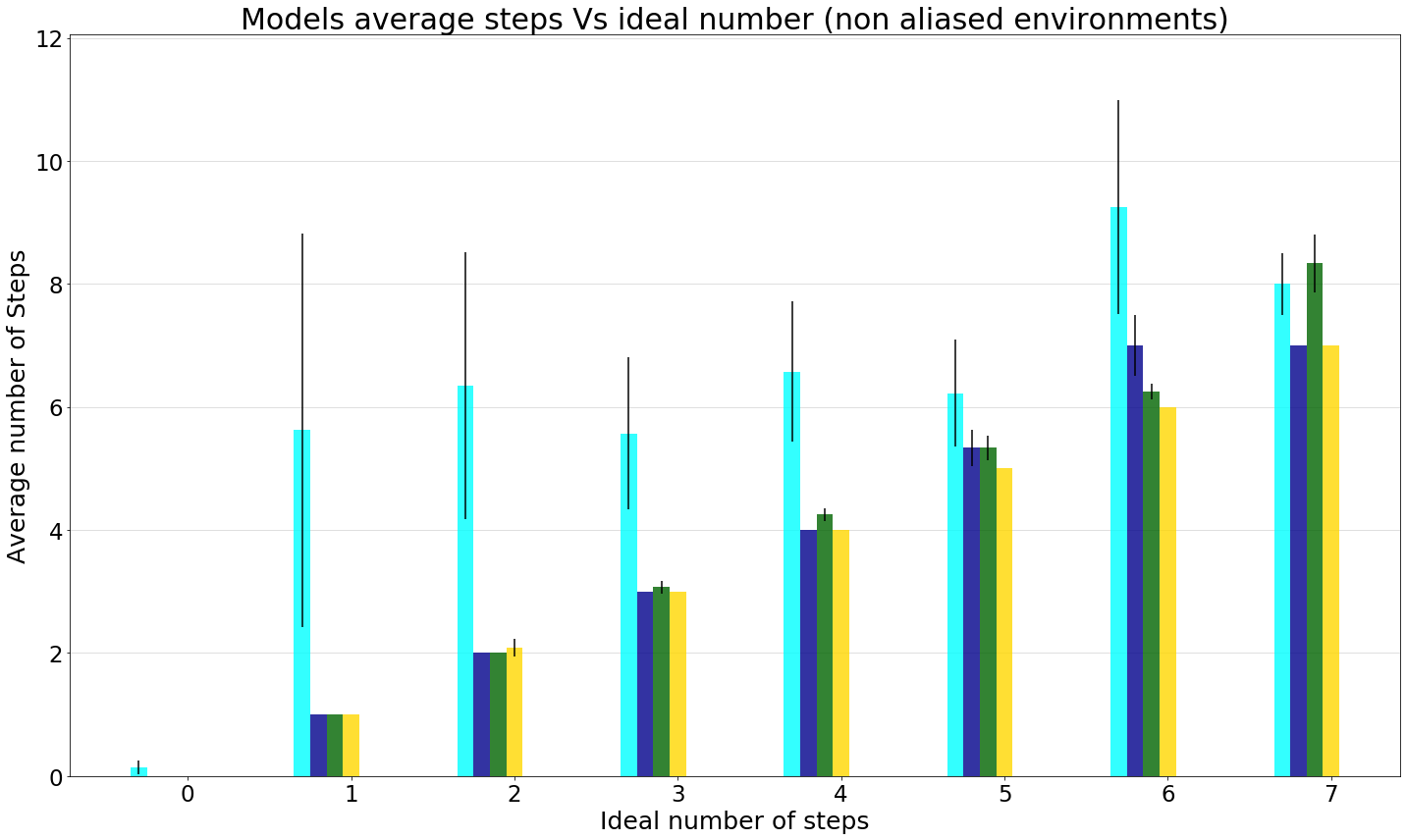}
        \caption{}
        \label{img:steps_goal_non_alias}
    \end{minipage}

\setcounter{figure}{7}
\setcounter{subfigure}{-1}
    \caption{Average number of steps needed by each model to reach the goal compared to Oracle's ideal steps. (a) Average steps over all aliased environments. (b) Average steps over all non-aliased environments. Knowing the environment allows the models to reach the goal more efficiently and closer to the ideal number of steps.}
    \label{img:steps_to_goal_all_models}
\end{subfigure}

While exploring, our agent prioritised information gain, in the following experiments we give a preferred observation (a floor colour) to stimulate exploitative behaviour instead. The agent can navigate guided by a preference regardless of whether the environment is familiar or not with the desire to reach that objective. A goal is considered reached if the agent decides to stay at the desired location, thus discarding random stumbling upon it as a successful attempt. Figure~\ref{img:steps_to_goal_all_models} shows the average number of steps each model requires to reach the goal, considering both aliased and non-aliased environments. The results are averaged over environment type (aliased maze or not), each model had 20 runs per goal distance and environment. Without prior knowledge about the environment, our model systematically explores the map until it encounters the goal, resulting in a higher number of steps than the oracle, which has precise knowledge of the goal location. However, when we relocate our agent after exploration, keeping the map in memory (i.e. ours wt prior), the results align closely with the oracle in non-aliased environments (Figure~\ref{img:steps_goal_non_alias}) and approach oracle performance in aliased environments (Figure~\ref{img:steps_goal_alias}), where the agent may need a few steps to localise itself while searching for the goal. A demonstration of our localisation process is presented in Figure~\ref{img:imagined_paths_to_goal}.
CSCG, on the other hand, exclusively operates with prior knowledge of the environment and employs a Viterbi algorithm for navigation~\cite{viterbi}. Despite having prior information similar to our agent, CSCG's performance is not consistently superior when provided only with colour observations. As our agent, CSCG also requires self-localisation based on multiple observations in such scenarios. Overall our agent demonstrates excellent efficiency in reaching and recognising the goal. Even without prior over the map, the agent does not return to already explored areas unless necessary, demonstrating a biologically plausible path planning. 

The impact of aliased observations over our model is demonstrated in Figures~\ref{img:imagined_paths_to_goal}. The left panel of each figure presents the agent (represented by an X) provided with a prior distribution over the map (agent's state value reported at the upper left of each tile) and the policies' expected free energy with a darker grey colour signifying higher preference to proceed toward that path.  The right panel depicts the agent's confidence in its location (i.e. state) given collected observations.
The agent starts at the bottom of the T-maze (Figure~\ref{img:step0}), observing only the present colour, which is observable in three different locations (here in states 0,1 and 7). The agent can't tell from just this single observation where it is yet. This results in divided confidence between those three states when inferring its potential localisation. In the right panel, we can see that the agent considers going forward as the best option. Notably, the imagined T turns are incorrect due to the agent's confidence in its localisation being mainly split between states 0 and 1, while the agent is, in reality, at state 7.
Moving to Figure~\ref{img:step1}, representing step 1, the agent adjusts its internal beliefs regarding localisation based on the colour of the previous room and the current observation. This correction in beliefs leads to a refinement in the agent's perceived localisation, as can be seen in the associated bar plot.
In Figure~\ref{img:step2}, corresponding to step 2, the agent exhibits a significantly higher level of certainty regarding its whereabouts in location 1. It confidently determines that the goal is located either to the left or right but rules out the possibility of it being behind.
Finally, Figure~\ref{img:step3} portrays the fourth step, where the agent demonstrates full confidence in its localisation and successfully identifies the objective with dark grey shading on the goals. This high level of confidence indicates the agent's strong belief in its internal representations. Therefore, the next steps will correctly lead the agent either right or left toward the preferred observation.

\setcounter{figure}{8}
\setcounter{subfigure}{0}
\begin{subfigure}
\setcounter{figure}{8}
\setcounter{subfigure}{0}
    \centering
    \begin{minipage}[b]{0.49\textwidth}
        \includegraphics[width=\linewidth]{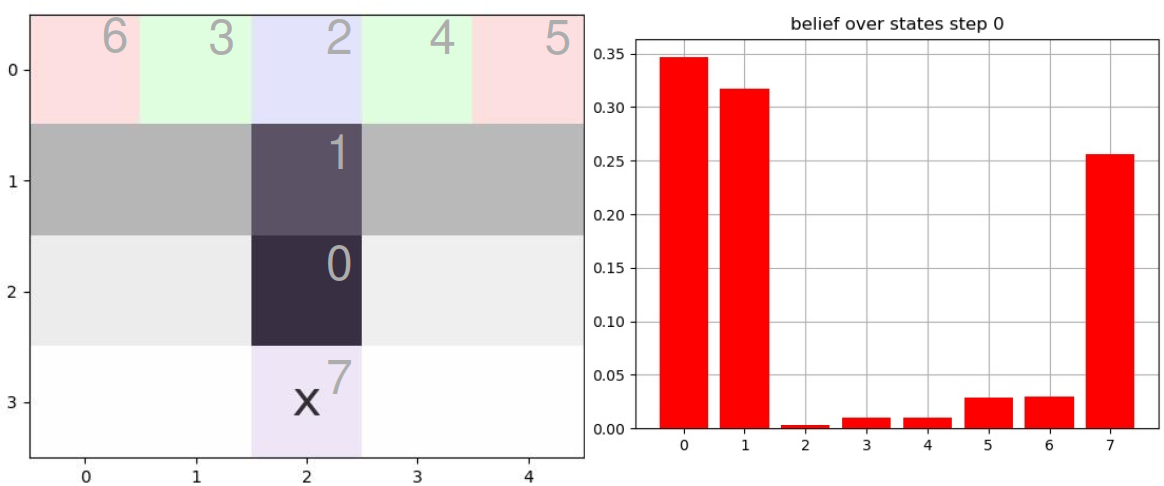}
        \caption{}
        \label{img:step0}
    \end{minipage}  
\setcounter{figure}{8}
\setcounter{subfigure}{1}
    \begin{minipage}[b]{0.49\textwidth}
        \includegraphics[width=\linewidth]{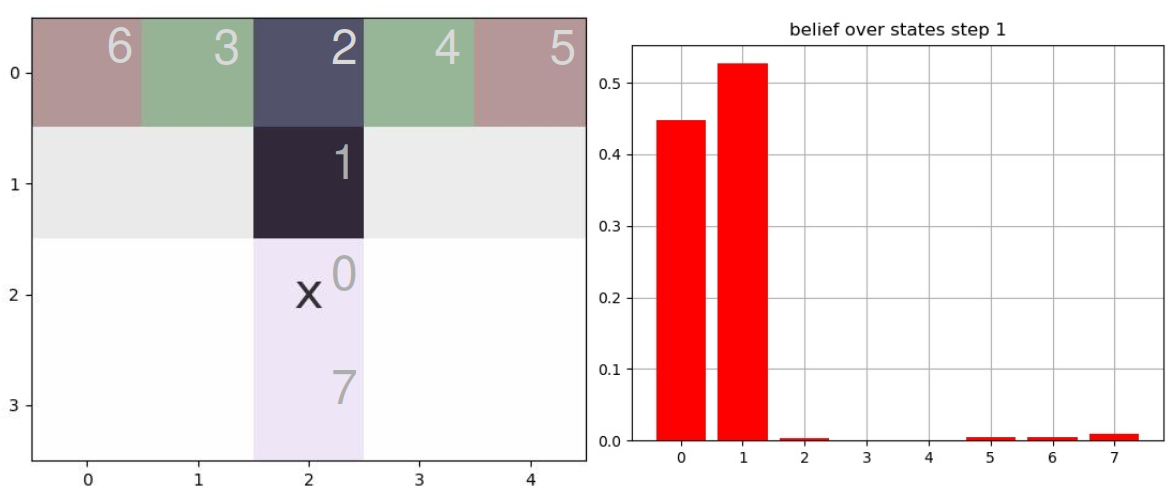}
        \caption{}
        \label{img:step1}
    \end{minipage}
    \setcounter{figure}{8}
\setcounter{subfigure}{0}

\setcounter{figure}{8}
\setcounter{subfigure}{2}
    \centering
    \begin{minipage}[b]{0.49\textwidth}
        \includegraphics[width=\linewidth]{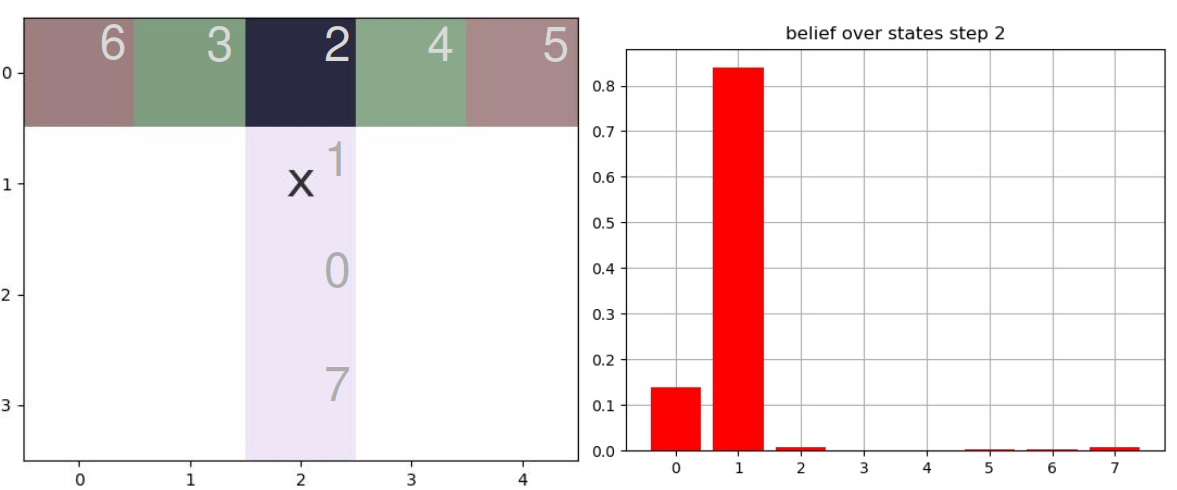}
        \caption{}
        \label{img:step2}
    \end{minipage}  
\setcounter{figure}{8}
\setcounter{subfigure}{3}
    \begin{minipage}[b]{0.49\textwidth}
        \includegraphics[width=\linewidth]{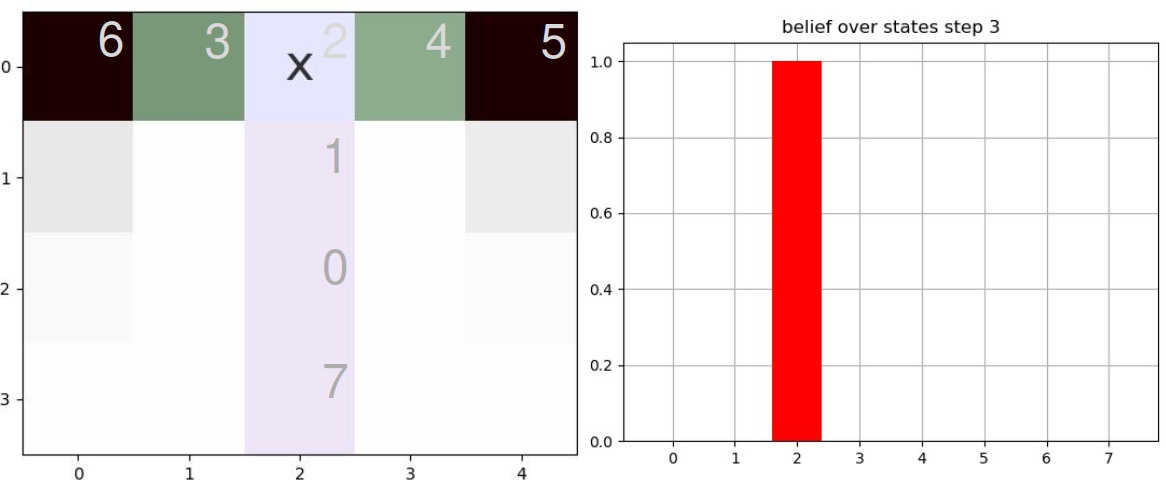}
        \caption{}
        \label{img:step3}
    \end{minipage}
\setcounter{figure}{8}
\setcounter{subfigure}{-1}
    \caption{Our agent imagined trajectories (left panel) alongside their corresponding localisation beliefs (depicted as red bars) at each step. Each number within the aliased T-maze signifies the internal state attributed by the agent during exploration. The agent's position is marked by an X, while the goal is a red observation at the end of the aisles. The termination points of the T aisles and the shading of the grey path reflect the agent's confidence in its beliefs.}
    \label{img:imagined_paths_to_goal}
\end{subfigure}

      

\FloatBarrier
\subsection{Dynamic Remapping}

Can the model correctly update its internal cognitive map given new evidence contradicting its prior? 
To verify that we conduct two experiments. One in the Donut environment, where there are only two paths to reach a goal (a long and a shorter path), by going left or right and a second one reproducing Tolman's second maze experiment with obstacles located at several positions.

\begin{figure}[!htb]
    \centering
    \includegraphics[width=15cm]{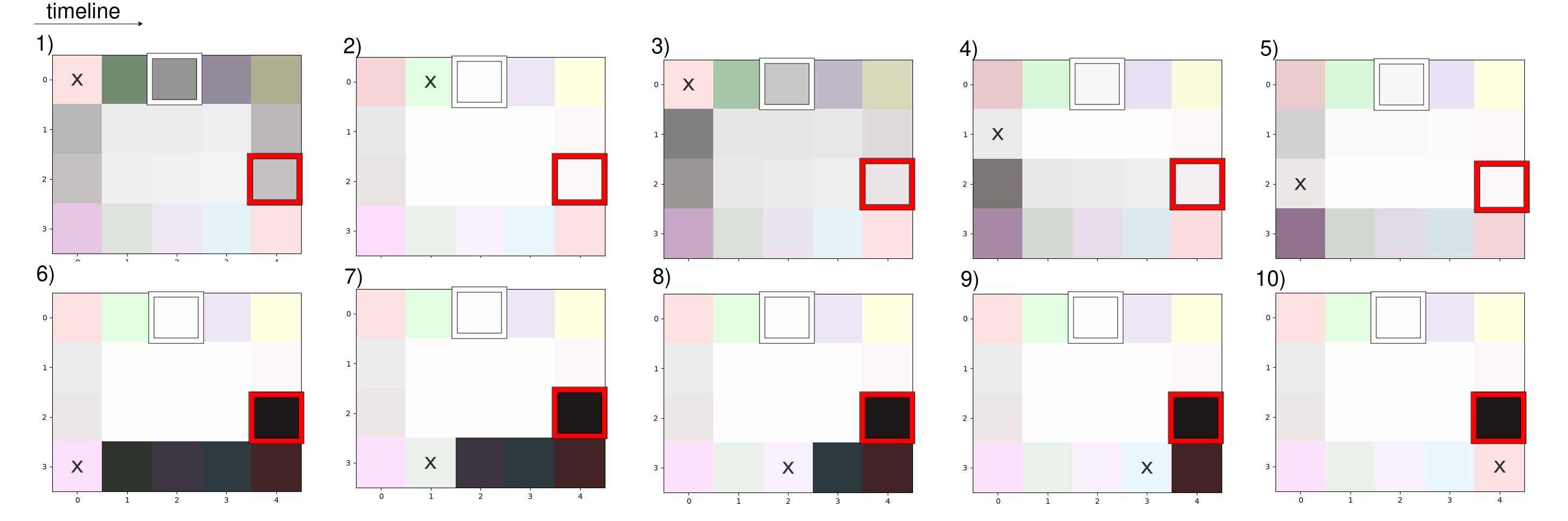}
    \caption{Our agent imagined path toward the objective (squared red) and re-planning when realising the desired path is blocked (closed room squared white).}
    \label{img:obstacle_toward_goal}
    \vspace{-0mm}
\end{figure}

In our Donut experiment, the agent has previously learned the environment from a randomly selected starting point without any obstacle. Then, the agent was kidnapped to a corner of the map with the newly implemented objective of observing a lightly pink colour (highlighted by a red square in Figure~\ref{img:obstacle_toward_goal}). Our model is given a floor colour as objective and exploitation and exploration have equivalent weights in the navigation. Path planning steps should be followed following the sequence from 1) to 10) depicted in Figure~\ref{img:obstacle_toward_goal}. At first, the agent successfully plans its path to accommodate the new objective by moving through the upper path, as depicted in the first frame 1) of Figure~\ref{img:obstacle_toward_goal}. The dark grey shading indicates the high level of confidence in this path, the darker the shade, the higher the certitude this imagined policy leads to a desired outcome. We disrupted the shortest path by adding an obstacle, as highlighted by a white square in Figure~\ref{img:obstacle_toward_goal}. This obstruction invalidated the agent's original route, leading to a notable decrease in the probability of reaching the goal along the intended path (as evident in frame 2) of Figure~\ref{img:obstacle_toward_goal}). Consequently, the agent initiates a remapping process promptly, updating its beliefs regarding graph connectivity. The subsequent frames, from frames 3) to 10) in Figure~\ref{img:obstacle_toward_goal}, depict the agent's increasing confidence as it navigates closer to the objective using the longest path.

 In this scenario where a room along the imagined path was unexpectedly closed off, the agent demonstrates its capability to dynamically adapt its navigation strategy in response to changes in the environment, effectively leveraging new observations to revise its path and achieve its objective. Failure to imagine an alternative path would have resulted in the agent stubbornly trying to pass a closed door. This flexible connectivity exists despite using a generative model, which is known to have a hard time revising strong beliefs. This is due to two things, firstly, when experimenting with a blockage, our agent updates its internal model with negative parameter learning (see appendix~\ref{app:models_params}). Secondly, the inherent growth of our agent, as it grows in its state or observation dimension, the probabilities get more distributed, so transitions to past states that have not been experimented for a long time naturally weaken over time. Reproducing the mechanism of animal-like synaptic plasticity~\cite{hippo_cognitive_map}.

Our second test to assert the robustness of our model to dynamic environments entails replicating the Tolman maze experiment outlined in \cite{Tolman_maze}. This maze configuration, illustrated in Figure~\ref{img:tolman_maze}, involves placing starving rats at a starting point (marked by a red triangle) with food positioned at the opposite end of the map. Three routes are available to reach the food, with Route 1 being the shortest and Route 3 the longest. However, these routes can be blocked at points A and B, where the blockages affect the accessibility of Routes 1 and 2.
In our case, the agent can't observe obstacles from a distance, it has to be in contact with them from an adjacent room to learn about it. Each coloured grid in Figure~\ref{img:agent_path_wt_obs} represents an independent closed room and the white box represents an obstacle disposed at positions A and B. Thus, we rely on the "insight"~\cite{insight} of our agent and its ability to update its beliefs rather than its perceptual capabilities to reach the goal.

This "insight" of our agent is based on two key parameters: its planning ability, determining how far ahead it can imagine, and cognitive map plasticity, dictating how well it can adjust past beliefs to new evidence. We allowed the agent to predict action consequences up to 14 steps ahead, thus from any obstacle, the agent could imagine reaching the goal anyway. 

The model is provided with a red visual observation as its preferred prior throughout the entire experiment with a weight on EFE utility term of 2 (which results in the agent favouring reaching the objective over exploring). The 10 agents consistently start at the bottom of the maze as depicted in figure~\ref{img:tolman_maze} red triangle. The agents followed three series of 12 runs in the Tolman maze with no obstacle (figure~\ref{img:agent_path_wt_obs}.o), then with an obstacle at position A (figure~\ref{img:agent_path_wt_obs}.A) and finally at position B (figure~\ref{img:agent_path_wt_obs}.B).

\begin{subfigure}
    \centering
    \begin{minipage}[b]{0.25\textwidth}
        \includegraphics[width=\linewidth]{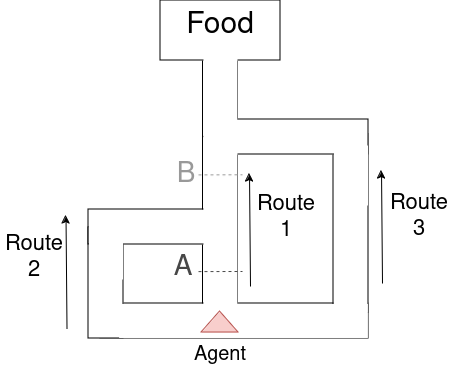}
        \caption{}
        \label{img:tolman_maze}
    \end{minipage}  
    \begin{minipage}[b]{0.60\textwidth}
        \includegraphics[width=\linewidth]{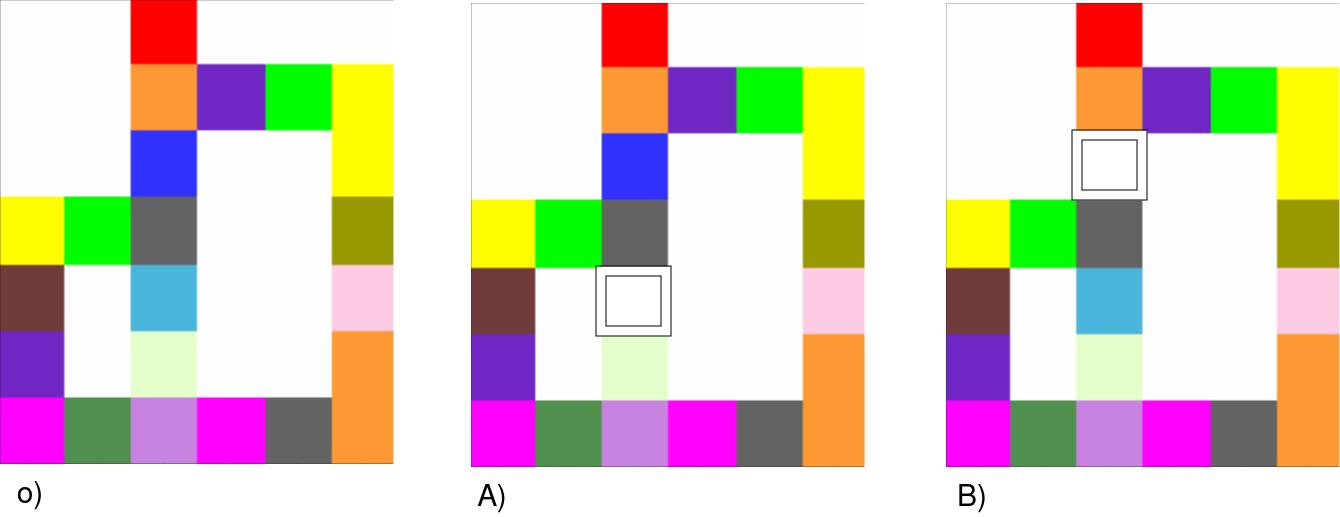}
        \caption{}
        \label{img:agent_path_wt_obs}
    \end{minipage}

\setcounter{subfigure}{-1}
    \caption{(a) Floor plan with obstacle location inspired by Tolman and Honzik's second maze \cite{Tolman_maze}, (b) shows the actual observation map without and with the blockage on position A) and B) }
\end{subfigure}

\begin{figure}[!htb]
    \centering
    \includegraphics[width=14cm]{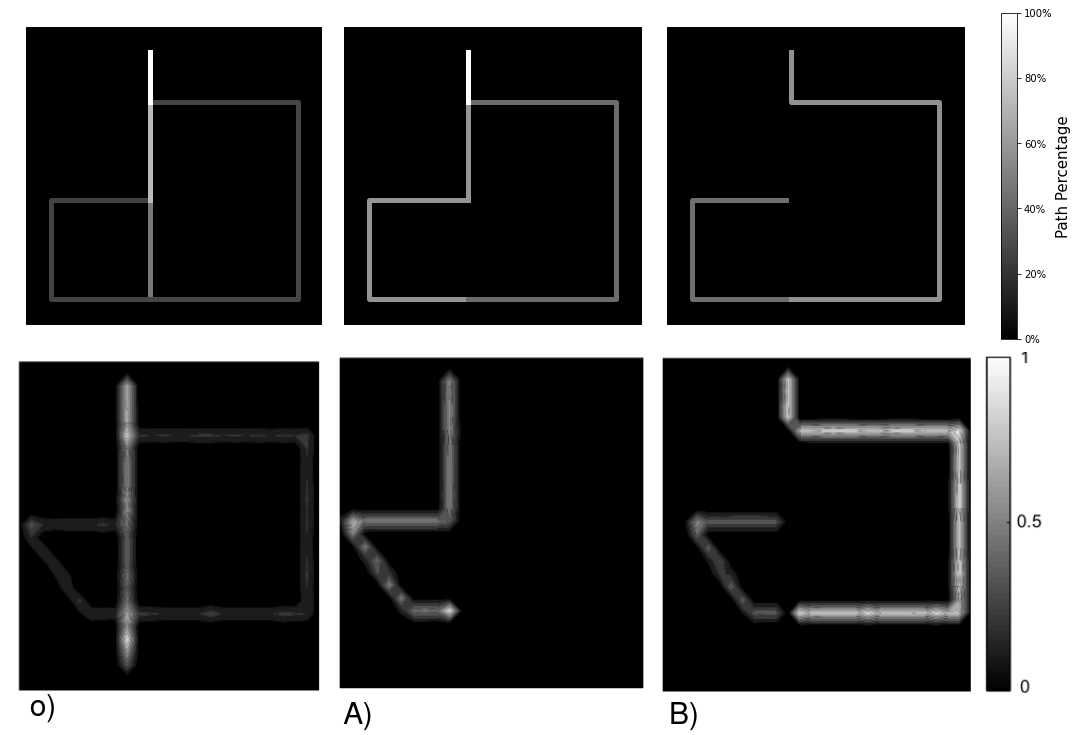}
    \caption{Our results compared to L.-E. Martinet \& al's~\cite{tolman_maze2_redone}. In our study, the agent's flow paths towards the objective (top of the map) are shown, with re-planning occurring when the desired path is blocked. The varying colour gradient of the lines indicates the frequency of selection for each path over all agents. Panels A and B illustrate obstacles at points A and B, respectively. The second row is adapted from L.-E. Martinet \& al's work~\cite{tolman_maze2_redone}. The occupancy grid maps demonstrate the learning of maze topology by simulated animals, initially without obstacles, showing a significant preference for Route 1. When a block is introduced at point A, the animals predominantly choose Route 2. With an obstacle placed at point B, the animals mainly opt for Route 3.}
    \label{img:martinet}
    \vspace{-0mm}
\end{figure}

Initially, the agent has no prior knowledge of the environment and observations. Every 20 steps, the agent is "kidnapped" and repositioned at the starting point without being informed of this relocation. The agent must infer and correct its belief about its location based on subsequent observations. 
Figure~\ref{img:martinet} shows the path choice frequency over all agents given an obstacle compared to ~\citet{tolman_maze2_redone2} results on 100 animats reproducing Tolman's insights. 

During the first 12 runs, the maze contains no obstacles, and the fastest route to the goal is path 1. Our agent typically explores all available paths at least once, following AIF framework as understanding the environment helps minimise free energy. Although the agent alternates between the three paths to balance between exploration and exploitative behaviour, path 1 is generally preferred (almost 50\% of the time) as we can see in the path count table~\ref{tab:path_count}, recording all the attempts to reach the goal after experimenting with the blockage in path 1, if any. It closely aligns with what we would expect from a hungry rat as shown in figure~\ref{img:martinet} first column, it resembles Tolman's expectations of rat behaviour. Our agent tends to alternate between taking the quickest path and taking another alternative path, with a clear regularity. Extending the number of runs would show path 1 being taken ~50\% of the time and paths 2 and 3 25\% of the time each. The reason why it isn't exactly 50\% of the time is because the agent takes a few runs to explore the environment instead of always reaching the goal. This alternate path selection is linked to the balance chosen between exploitation (the preference of being at the goal) and exploration (the preference of learning the environment). When a path is considered better understood than the others, the agent counterbalances by re-exploring those other paths. Let's remember that the agent can predict reaching the goal from those paths as well, even if it takes more steps. 

An obstacle is introduced at position A for the next series of 12 runs. The agent continues to navigate using the same memory, which has accumulated experience from the previous runs in an obstacle-free environment. Path 2 becomes the quickest route. After encountering the obstacle a few times, the agent updates its internal map and shifts its preference to path 2, though it still periodically checks paths 1 and 3, and reaffirms that path 1 remains blocked. Trials consider all the tentative to reach the goal after experimenting with the blockage in path 1. 
Contrary to~\citet{tolman_maze2_redone2} results, our agents show a more divided comportment with path 2 being preferred but not completely neglecting path 3.  
Finally, the obstacle is moved to position B, and the experiment continues for 12 more runs. Initially, the agents often attempt to traverse path 2, but after further updating its model to account for the new blockage, they predominantly switch to path 3, while periodically checking path 2 again.

These results demonstrate our agent's robustness in handling kidnapping scenarios and its ability to update its internal cognitive map when new evidence contradicts previously well-verified beliefs. The model exhibits a clear and adaptable behaviour, akin to rats navigating a maze. By adjusting the agent's prediction horizon, the weighting of the utility term in the Expected Free Energy, and the stochasticity in path selection (testing values defined in appendix~\ref{app:models_params}), we can control the navigation behaviour of our agent, favouring exploration or exploitation. These variations in behaviour are easily interpretable through the AIF framework our model uses, avoiding the opacity often associated with black-box models.

\begin{table}[h!]
\centering
\caption{Path count over all ten agents when they reach the goal considering the obstacle position. Trials, where the objective is not reached under 20 steps (exploration phases), are not counted.}
\begin{tabular}{|c|c|c|c|}
\hline
 & Path 1 & Path 2 & Path 3 \\ \hline
No obstacle & 48 & 27 & 28 \\ \hline
Obstacle A  & 0  & 70 & 49 \\ \hline
Obstacle B  & 0  & 36 & 48 \\ \hline
\end{tabular}

\label{tab:path_count}
\end{table}

\FloatBarrier

\section{Discussion}

This study proposes a novel high-level abstraction model grounded in biologically inspired principles, aiming to replicate key aspects of animal navigation strategies~\cite{Human_rodent_spatial_rep, humans-hierarchic-plan-subway}. Integrating a dynamic cognitive graph alongside internal positioning within an Active Inference Framework is central to our approach. This novel combination enables our model to dynamically expand its cognitive map upon prediction while navigating any ambiguous mini-grid environment with or without prior, mirroring the adaptive learning and efficient exploration abilities observed in animals~\cite{few_one_shot_learning,mice_in_labyrith}. Comparative experiments with the Clone-Structured Graph (CSCG) model~\cite{cscg_pres} underscore the superiority of our approach in learning environment structures with minimal data and without preparatory knowledge of specific observation and state-space dimensions. Moreover, we demonstrated the model's ability to adapt its cognitive map to new evidence and reach preferred observations in various situations. The presented model can represent large or complex environments like warehouses or houses, without significant memory or storage demands for the cognitive map itself due to its matrix-based structure. However, as the agent attempts to make long-term predictions—projecting further steps into the future—computational requirements for processing power and memory increase. In this work, we implement up to 13-step predictions without issue due to the efficiency of our policy design; achieving predictions beyond 10 steps is notable for this type of generative model. Looking ahead, it could be interesting to investigate the impact of a perfect memory on future policies and exploration efficiency in addition to measuring the impact of belief certitude on the graph dynamic adaptation. Furthermore, splitting the data by unsupervised clustering~\cite{self_labelling}, or by using the model's prediction error to chunk the observations into separate locations~\cite{chunking} would offer a closer approximation to animal behaviour and allow open space exploration. This could extend this model capacity into real-world scenarios, such as StreetLearn~\cite{street_learn} based on Google map observations or simulated realistic environments such as Habitat~\cite{habitat}. The real world presents additional challenges, such as segmenting large open spaces into manageable information chunks as realised by~\citet{explo_chunk}, efficiently processing visual data or other sensory inputs (with a method such as~\citet{Gaussian_splatting}), and distinguishing new observations from known ones based on memory. Moreover, navigating around local static and dynamic obstacles effectively is essential,~\citet{pred_control} propose a possible solution to move between locations while avoiding obstacles. Each of these challenges can be addressed individually and integrate the mentioned solutions to enhance our model’s performance without altering the core navigation strategy. Finally developing this high-level abstraction model with a hierarchical model~\cite{ours_2024} would allow the model to reason over different levels of abstraction temporally and spatially extending its navigation to local in-room navigation and long-term planning.





\section*{Funding}
This research received funding from the Flemish Government under the “Onder-zoeksprogramma Artificiële Intelligentie (AI) Vlaanderen” programme.


\section*{Data Availability Statement}
The code generated for this study can be found in the high-level nav planning: \url{https://github.com/my-name-is-D/high_level_nav_planning/tree/main}.


\bibliographystyle{Frontiers-Vancouver} 
\bibliography{main}

\FloatBarrier
\section*{Appendix}
\appendix
\section{Environments}
\label{app:envs}

\begin{figure}[!h]
    \centering
    \includegraphics[width=12cm]{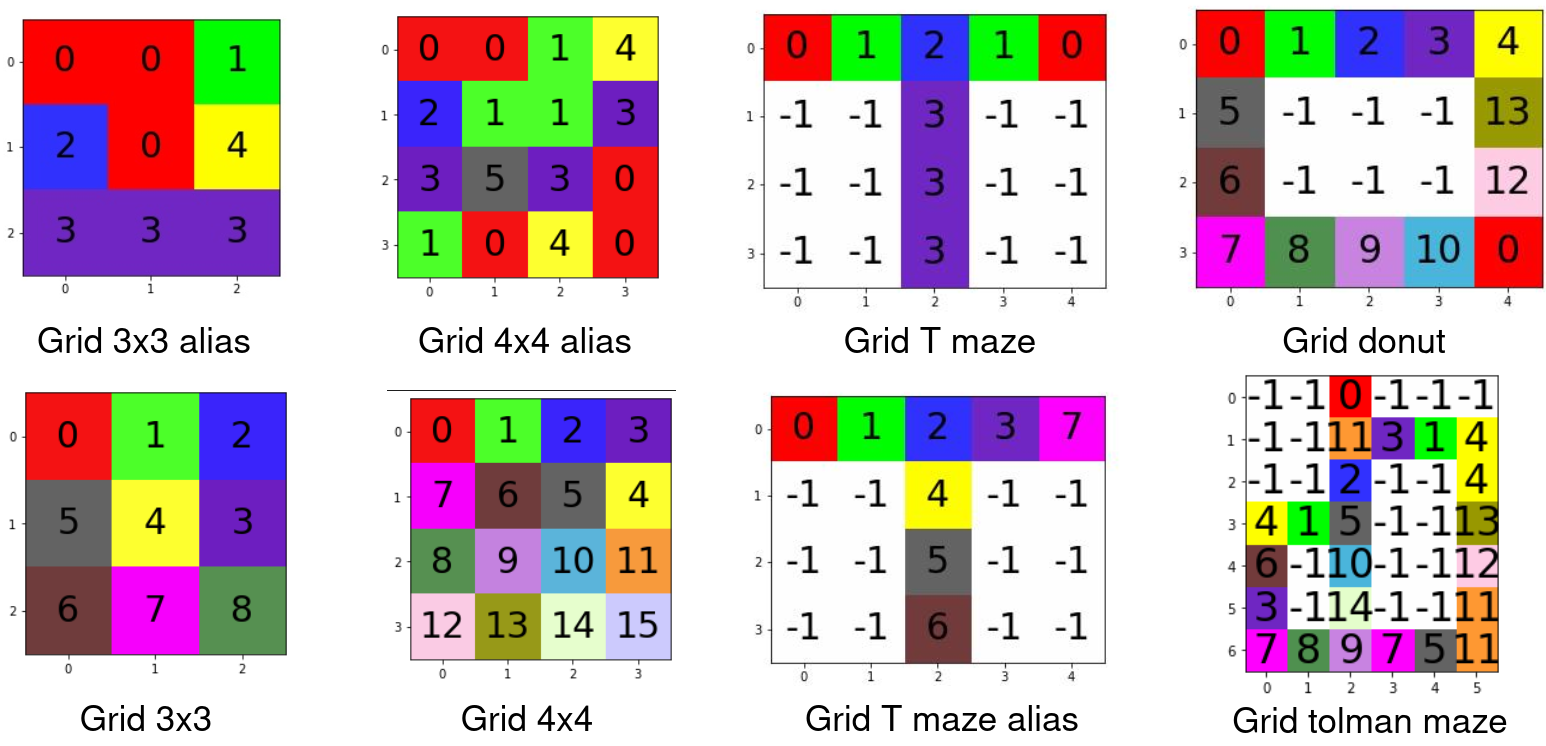}
    \caption{Environments observations with their names as used in this paper. The negative observations are un-observable by the agent.}
    \label{img:envs}
    \vspace{-0mm}
\end{figure}

The agent's perceptual capabilities are limited to perceiving the current room's colour and obstacle position. The colour observations received by the model are detailed in Figure~\ref{img:envs}, they are associated with numerical values to aid visualisation. Negative values correspond to un-observable positions, indicating that moving in that direction leads to an obstacle. Ground-truth positions are aligned with the axes of the environments, where each pose is represented as a (row, column) pair -(y, x)- within each environment.

\FloatBarrier

\section{Models complementary information}

Both models know in advance the actions they can take, defined in our mini-grid environments as being up, down, left, right and stay -do not move-.
To ease learning we inform both agents that staying means not changing state.

\subsection{Our Model Parameters}
\label{app:models_params}

Our model leverages the pymdp framework~\cite{pymdp}, a Python library designed for active inference in discrete state spaces. Initially, the model starts with a 2x2 dimension for both state and observation matrices due to Python's matrix requirements. However, only the first dimension contains meaningful information at the start (indicating the initial position and current observation). As the agent explores, the model dynamically expands its state and observation dimensions, incorporating new information from both its predictions and actual observations.

The agent contains as sub-models: 
\begin{itemize}
    \item the Markov matrices $A_o$ (observation likelihood) $P(o_t|s_t)$, $B_s$ (state transition) $P(s_t|s_{t-1}, a_{t-1})$ and $A_p$ (position likelihood) $P(p_t|s_t)$. 
    \item  the list of successive predicted poses as tuples $B_p$ (position transition) $P(p_t|p_{t-1},a_{t-1})$
\end{itemize}


The transition probability update, as depicted in Equation \ref{eq:B_up}, incorporates various learning rates based on different scenarios. Specifically, an experimented transition employs a learning rate of 10, while the reverse transition is assigned a rate of 7. An experimented impossible transition (state can't be reached because of an obstacle) is given the same values but a negative learning rate. In contrast, a predicted transition adopts a learning rate of 5 in the forward direction and 3 in the reverse direction. This difference in learning rate is used to adjust the certainty we have about the chosen policy $\pi$. Table~\ref{tab:tran_lr} recapitulates those values.

\begin{table}[h!]
\centering
\caption{Transition learning rate depending on the situation}
\begin{tabular}{lccc}
\toprule
\textbf{Transitions} & \textbf{Possible} & \textbf{Impossible} & \textbf{Imagined} \\
\midrule
Forward  & 7  & -7 & 5 \\
Reverse  & 5  & -5 & 3 \\
\bottomrule
\end{tabular}

\label{tab:tran_lr}
\end{table}

A higher learning rate reflects more confidence in the policy, while a lower learning rate indicates uncertainty. By weighting the learning rate, the model ensures that the transition probability is considered in its decision-making in dynamic environments.

\begin{equation}
    B_\pi = Q(s_t|s_{t-1}, \pi) Q(s_{t-1}) * B_\pi  * learning\_rate
    \label{eq:B_up}
\end{equation}

The likelihood probability update is similar with a learning rate fixed to 1.

To assess the agent's confidence in its current pose, we evaluate the certainty associated with its inferred state by considering both observation and pose. If this confidence falls below 70\%, the position is disregarded for state inference until the agent can achieve a clearer understanding of its whereabouts. This ensures that the model relies only on well-supported information when making decisions, thereby improving the accuracy and reliability of its state estimations.

The expected free energy is calculated for each future time-step as defined in equation~\ref{eq:planning_as_inf} the agent considers and is then aggregated to infer the most likely sequence of actions to reach a preferred state. This belief in policies is achieved through:
\begin{equation}
    P(\pi) = \sigma (-\gamma G(\pi))
\end{equation}

Where $\sigma$, the softmax function is tempered with a temperature parameter $\gamma$ converting the expected free energy of policies into a categorical distribution over policies. We set $\gamma$ to 16 in our model to add more weight to favour policies minimising EFE. 

\subsection{Our Model Policies Generation}
\label{app:policies}
We have tried diverse policy creation to reduce the computational load on imagining all possible transitions. Policy algorithms are linked to our models' versioning as well. We have 3 different ways to create policies, from the most computationally expensive to the lightest.

Considering the look-ahead as being the number of steps the agent can project itself over. The first policy creation is simply implied to consider all possible combinations for our 5 actions (Up/Down/Left/Right/Stay). 
This results in an exponential number of combinations: $n\_policies = 5^{lookahead}$. 
Our second policy generation method creates paths covering all possible motions in the look-ahead range while making sure a path doesn't return on its track. The 'STAY' action is added after every new action as an independent path, such that the agent can imagine stopping at any time. This results in the approximated parabolic equation:
\begin{equation}
    n\_policies   \tilde{=}   86.3*lookahead^2 - 392.0*lookahead + 377.0
\end{equation}

Finally, the lighter policy generation produces simple L-shaped paths covering the look-ahead range \cite{ours_2024}, adding the 'STAY' action at each new action added in a path as a new independent policy. The complexity of this solution is polynomial :
\begin{equation}
    n\_policies = 4*lookahead^2 +1
\end{equation}

Without the STAY action, it would become linear, allowing for a very large look-ahead without worrying about computation cost. However, the model accuracy on imagined consequences over a long prediction range is also to consider and has not been analysed in this study.   

Whichever policy generation strategy is used all the generated policies follow the AIF process of selection described in~\cite{nav_aif}.  

In this work, we use 16 as a constant for the temperature $\gamma$. This means that we are highly likely to choose the optimal policy rather than any other one. 

\subsection{CSCG Model Parameters}
\label{app:cscg}
The CSCG model is imagined as a static matrix, it is therefore given the information of how many observations it is going to encounter at initialisation. The number of clones has been set up to 10 so it could work in all of our environments without issue. This means that the CSCG model can't cope with environments larger than what its model dimension can incorporate.

While training its internal beliefs given the sequence of actions and observations we set the pseudocount to 0.05 and let the agent train for 200 iterations, then for the Viterbi optimisation of the states we set the pseudocount to 0.0001 and let it run for 100 iterations.
The pseudocount is a small constant ensuring that any transition under any action has a non-zero probability. It also improves convergence, however, too small during training and it gives wrong state estimations. All values have been carefully selected to give the best result possible in the smallest amount of recursion. More details relative to the model can be found in~\cite{cscg_pres}. In this study, we re-train the CSCG model for each 5 consecutive steps of the exploration, considering all the past observations and actions at each re-train.


\subsection{Computational Costs}

We averaged the computation time over 5 runs for both our model and the CSCG model to complete 100 exploration steps.
The results indicate that our model is significantly faster, underscoring its relevance for real-world applications. The slower performance of the CSCG model can be attributed to its need to fully retrain every 5 steps, whereas our model continuously updates its internal map with predicted and observed changes at each step. In these tests, our model operated with a look-ahead prediction range of 6 steps.
\begin{table}[h!b]
\centering
\caption{Computational time in seconds of ours and CSCG model while forced to explore during 100 steps in an environment. Our model imagines the policies over 6 consecutive steps. 
}
{
\begin{tabular}{lll}
model & Ours& CSCG \\ \hline
\begin{tabular}[c]{@{}l@{}}Execution\\ time (s)\end{tabular} & 14.95 & 48.35 \\ \hline
\end{tabular}
}
\end{table}

\FloatBarrier

\end{document}